\documentclass{article}
\usepackage[preprint]{neurips_2026}

\usepackage[utf8]{inputenc}
\usepackage[T1]{fontenc}
\usepackage{hyperref}
\usepackage{url}
\usepackage{booktabs}
\usepackage{amsfonts}
\usepackage{amsmath,amssymb,amsthm}
\usepackage{microtype}
\usepackage{xcolor}
\usepackage{graphicx}
\usepackage{multicol}
\usepackage{multirow}
\usepackage{enumitem}
\usepackage{algorithmicx}
\usepackage{algpseudocode}
\usepackage{subcaption}
\usepackage{caption}
\usepackage{wrapfig}
\usepackage[most]{tcolorbox}
\usepackage{xcolor}
\newtcolorbox{promptbox}[1]{
  colback=gray!5,
  colframe=gray!60,
  boxrule=0.5pt,
  arc=2pt,
  left=6pt,
  right=6pt,
  top=5pt,
  bottom=5pt,
  fonttitle=\bfseries,
  title=#1,
  breakable
}

\usepackage{algorithm}
\usepackage{algpseudocode}

\algrenewcommand\algorithmicrequire{\textbf{Input:}}
\algrenewcommand\algorithmicensure{\textbf{Output:}}
\algrenewcommand\algorithmiccomment[1]{\hfill$\triangleright$ #1}

\newcommand{\method}[1]{\textsc{#1}}
\newcommand{\model}{\method{CoEvoT}{}}
\renewcommand{\algorithmicrequire}{\textbf{Input:}}
\renewcommand{\algorithmicensure}{\textbf{Output:}}
\newcommand{\stitle}[1]{\vspace{1mm} \noindent {\bf #1}}
\renewcommand{\vec}[1]{\ensuremath{\mathbf{#1}}}

\title{\model: Co-Evolving Chain-of-Thought Prompting for Graph--LLM Reasoning}

\author{%
\textbf{Haohua Niu\textsuperscript{1}
\quad Xingtong Yu\textsuperscript{2*}
\quad Yang Liu\textsuperscript{3}
\quad Junfeng Fang\textsuperscript{4}}
\quad {Xuanting Xie\textsuperscript{5}}  \\
\textbf{Jie Tan\textsuperscript{2}
\quad Zhongjian Zhang\textsuperscript{6}
\quad Hong Cheng\textsuperscript{2}
\quad Yuan Fang\textsuperscript{7}} \\
\textsuperscript{1}Sun Yat-Sen University
\quad \textsuperscript{2}The Chinese University of Hong Kong \\
\textsuperscript{3}Institute of Computing Technology, Chinese Academy of Sciences \\
\textsuperscript{4}National University of Singapore \quad
\textsuperscript{5}University of Electronic Science and Technology of China \\
\textsuperscript{6}Beijing University of Posts and Telecommunieations \quad
\textsuperscript{7}Singapore Management University\\
}

\begin{document}

\maketitle

\begin{abstract}
Graph learning under distribution shift presents a persistent challenge, where models adapt to new graphs with limited or even no supervision. Recent graph--LLM approaches move toward label-efficient prediction by linearizing graphs into prompts and using large language models (LLMs) as predictors, and can adopt Chain-of-Thought (CoT) prompting to exploit LLM's multi-step reasoning capability. However, existing CoT-based graph--LLM methods generate intermediate thoughts while conditioning on fixed graph tokens, limiting step-wise refinement of structural cues.
In this paper, we propose \model, a simple yet effective co-evolving CoT prompting framework for graph--LLM reasoning. \model\ couples \emph{text-to-graph} token rewriting and \emph{graph-to-text} reasoning guidance in a closed loop: each intermediate textual thought is used to update the graph token evidence state via a lightweight condition network, and the updated tokens are fed back into the next-step instruction to guide subsequent LLM reasoning. This enables step-wise, state-aware evidence refinement, rather than reasoning over a fixed graph snapshot.
Extensive experiments on eight datasets demonstrate that \model\ consistently outperforms state-of-the-art baselines.
\end{abstract}

\section{Introduction}
Graphs provide a compact abstraction for relational data, supporting applications across citation analysis~\citep{10.1145/3711896.3737031,yu2025gcot}, social networks~\citep{qiu2020gcc,10.1145/3690624.3709192}, e-commerce~\citep{10.1145/3580305.3599320,sun2022gppt}, and biological systems~\citep{10.1145/3534678.3539213,tran2025groot}. While supervised graph neural networks (GNNs)~\citep{kipf2016semi,velivckovic2017graph,hamilton2017inductive} are effective with sufficient labels, their performance often degrades under limited supervision, and they seldom transfer reliably to new graphs with shifted structure or semantics.  Self-supervised graph pre-training improves label efficiency by learning from unlabeled data and adapting via fine-tuning~\citep{velickovic2019deep,you2020graph}, yet fine-tuning can still be costly and sensitive to mismatches between pre-training and downstream objectives.  Graph prompting further reduces adaptation overhead by freezing the backbone and optimizing lightweight prompts~\citep{liu2023graphprompt,yu2024generalized,yu2024few}; however, it generally depends on labeled feedback to tune prompts and struggles to generalize across datasets without supervision.

\begin{wrapfigure}[18]{r}{0.5\columnwidth}
    \centering
    \includegraphics[width=\linewidth]{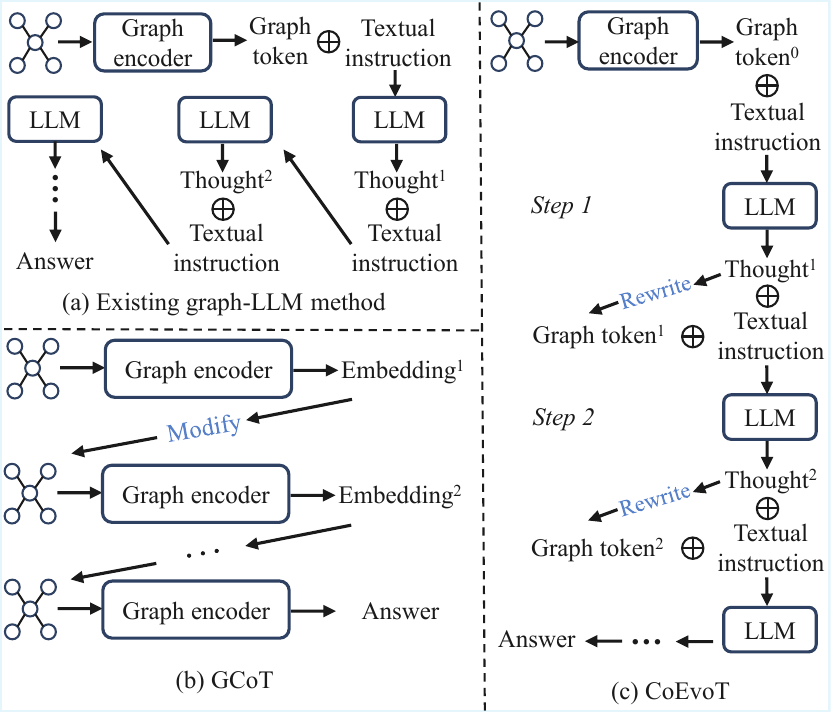}
    \vspace{-7mm}
    \caption{Comparison of existing graph--LLM methods, \method{GCoT} and the proposed \model.}
    \label{fig:motivation}
\end{wrapfigure}

The rapid progress of LLMs offers a new route to label-efficient graph learning.  Prevailing  approaches~\citep{chen2024llaga,tang2024graphgpt} linearize a graph into a sequence of tokens and combine it with textual instructions, guiding the LLM to produce predictions directly. Meanwhile, CoT prompting has emerged as an effective mechanism for improving LLM-based  reasoning~\citep{wei2022chain,wang2023self}.  Rather than answering in a single pass, CoT decomposes reasoning into multiple steps, generating intermediate thoughts that progressively lead to the final answer, as shown in Fig.~\ref{fig:motivation}(a). This step-wise mechanism is  appealing for graph-centric problems, where predictions often hinge on iterative evidence aggregation and reconciling local patterns with broader structural context~\citep{yu2025gcot}.

Although existing graph--LLM methods can incorporate CoT prompting~\citep{Jin2024GraphCA,chen2024llaga,tang2024graphgpt}, their reasoning remains largely language-only: intermediate thoughts are typically triggered by artificial instructions (e.g., ``think step by step''), while the graph tokens used to construct the LLM instruction is fixed throughout reasoning. This static token design limits what CoT can achieve on graphs, where effective prediction often requires progressively re-weighting neighborhoods and paths as the state evolves. In parallel, a recent CoT-style method for text-free graphs~\citep{yu2025gcot} enables step-wise refinement of graph embeddings, as shown in Fig.~\ref{fig:motivation}(b). However, its ``thoughts'' are latent embedding states rather than interpretable reasoning steps, and it does not exploit the rich textual attributes available in many real-world graphs.

Hence, CoT for graph--LLM reasoning should co-evolve in both language and structure, rather than treating either side as static context. In a multi-step reasoning, intermediate textual thoughts should serve as conditioning signals that update the structural representation, rewriting the graph tokens to reflect the current state. The updated graph tokens then act as more targeted evidence, guiding subsequent LLM reasoning with an updated state-aware view of the graph. This calls for a co-evolving CoT mechanism that tightly couples \emph{text-to-graph} token rewriting with \emph{graph-to-text} reasoning guidance in a closed feedback loop. 
In this work, we propose a simple yet effective \textbf{Co}-\textbf{Evo}lving Co\textbf{T} prompting framework (\model) for graph--LLM reasoning that tightly couples text-to-graph token rewriting with graph-to-text reasoning guidance. Rather than treating a graph-derived context as a one-off prompt, \model\ turns it into an explicit, step-wise evidence state that can be iteratively refined and re-consumed during reasoning.

First, \textit{to enable state-aware text-to-graph rewriting across multiple reasoning steps}, we introduce a thought-conditioned graph-token rewriting module. At each step, a lightweight condition network uses the intermediate thought as a conditioning signal to generate graph prompts that adaptively update the graph-token state for the next step. In this way, intermediate thoughts are no longer merely textual traces, they serve as actionable controls that continuously reshape the structural evidence.

Second, \textit{to provide graph-to-text guidance throughout multi-step reasoning}, we maintain the graph tokens as a persistent evidence state and inject the updated state into the next-step instruction, together with the task prompt and accumulated thoughts. This grounds each subsequent round of LLM reasoning in the \emph{latest} state-aware structural context, rather than a fixed snapshot, driving co-evolving reasoning across language and structure.

In summary, the contributions of this work are fourfold.
(1) We identify the limitation of CoT prompting in existing graph--LLM methods, which perform step-wise reasoning over static graph evidence. (2) To address the issue, we propose \model, a co-evolving CoT prompting framework that couples text-to-graph token rewriting with graph-to-text reasoning guidance for stronger cross-dataset transfer.
(3) Specifically, we introduce a thought-conditioned rewriting module to update graph tokens with intermediate thoughts, and feeds the updated tokens back into subsequent instructions to ground the LLM in  state-aware structural context.
(4) We conduct extensive experiments on eight datasets, demonstrating consistent improvements over strong state-of-the-art baselines.

\section{Related Work}
\stitle{Graph neural networks.}
GNNs~\citep{feng2023extending,kong2023mag,wang2024gnnboundary,pahng2024improving,zhao2025graph,bechler2309graph} are typically based on message passing~\citep{feng2022powerful,xu2018powerful,alon2020bottleneck}, where node representations are updated by aggregating neighborhood information. 
Despite strong empirical performance, most GNN pipelines are supervised and require task-specific retraining or fine-tuning, which limits scalability under label scarcity and distribution shift.
To improve transferability, self-supervised graph pre-training learns general-purpose representations from unlabeled graphs~\citep{velivckovic2018deep,qiu2020gcc,you2021graph,hou2022graphmae,he2025unigraph}. 
More recently, graph prompt learning adapts frozen pre-trained encoders via lightweight prompts~\citep{liu2023graphprompt,yu2024generalized,yu2024few}. While effective in low-resource settings, prompting still relies on task feedback to select or optimize prompts, and thus remains largely few-shot rather than zero-shot on unseen datasets.

\stitle{LLMs for graphs.}
A line of work uses LLMs as semantic enhancers for GNNs by generating auxiliary signals, such as task descriptions, contextual prompts to guide downstream learning~\citep{tian2024graph,hu2025large,zolnai2024stage,qiao2025login}.  While these methods benefit from the rich prior knowledge of LLMs, the final predictions are still produced by GNNs, constraining generalization due to inherent GNN inductive biases. Several studies further introduce architectural designs to enable zero-shot transfer across datasets \citep{qiao2025login,sun2022gppt,sun2023graph,fang2023universal}, yet their generalization remains limited in practice. 
Another direction treats LLMs as \emph{predictors} by converting graphs into text/sequences or discrete tokens for direct prompting-based inference~\citep{zhao2023graphtext,zhang2024graphtranslator,yu2025graph2text,chen2024llaga}. 
This paradigm is appealing for zero-shot transfer, since it can apply a unified instruction format without task-specific training. 
However, the graph evidence presented to the LLM is typically static during inference: the model reasons over a fixed representation of the graph, even when multi-step prompting such as CoT is used.  This static-evidence design limits iterative refinement, as the relevance of neighborhoods, paths, and relational cues should be progressively re-weighted as state evolve.

\section{Preliminaries}
\label{Preliminaries}
In this section, we present preliminaries and problem definition.

\stitle{Graph.} A graph is defined as $G=(V,E,\vec{X})$, where $V$ is the set of nodes and $E$ is the set of edges. The nodes are associated with a feature matrix $\vec{X}\in\mathbb{R}^{N\times d}$, where $\vec{x_i}\in\mathbb{R}^{d}$ is the feature of $v_i$.

\stitle{Graph encoder.} 
GNNs are typically built upon message passing, where node features are iteratively updated by aggregating information from local neighborhoods~\citep{scarselli2008graph}.
Formally, let $\vec{H}^l$ denote the node embedding matrix at the $l$-th layer, the message-passing is:
\begin{equation}
\vec{H}^l = \mathrm{MP}(\vec{H}^{l-1}, G; \boldsymbol{\theta}^l),
\end{equation}
where $\boldsymbol{\theta}^l$ denotes the learnable parameters.
The initial embeddings are given by $\vec{H}^0 = \vec{X}$.
After stacking $L$ layers, we obtain the final node embeddings
$\vec{H} = \vec{H}^L$.
For brevity, the multi-layer encoding process is summarized as
\begin{equation}\label{eq.gnn}
\{\vec{H}^1, \ldots, \vec{H}^L\}
= \mathrm{GE}(\vec{X}, G; \boldsymbol{\Theta}),
\end{equation}
where $\mathrm{GE}$ stands for the graph encoder, $\boldsymbol{\Theta} = \{\boldsymbol{\theta}^1, \ldots, \boldsymbol{\theta}^L\}$.

\stitle{Pre-training}. Existing studies on graph--text self-supervised learning show that contrastive pre-training can effectively align structural graph representations with textual semantics \citep{wen2023augmenting,zhu2025graphclip}; following prior work~\citep{wang2024llms}, we adopt a multi-modal pre-training strategy that jointly exploits graph structural signals and the semantic space of LLM. 
Specifically, given an input graph $G$, we construct two augmented views $G_1$ and $G_2$ by perturbing graph structure and node attributes. These two views are encoded by a shared graph encoder $\mathrm{GE}(\cdot)$ to produce paired graph representations. Representations of the same node across $G_1$ and $G_2$ form positive pairs to enforce cross-view graph consistency, while representations of other nodes within the batch are treated as negative samples.

To further incorporate language semantics, we introduce the token embedding space of the LLM, denoted as $\mathcal{E}_{\text{LLM}}$, as a semantic reference space. Graph representations derived from $G_1$ and $G_2$ are contrasted against $\mathcal{E}_{\text{LLM}}$, encouraging structurally corresponding representations to align with their semantic counterparts while remaining discriminative from others.
Formally, the pre-training objective can be abstracted as:
\begin{equation}
\mathcal{L}_{\text{pre}}(\Theta)
=
\mathcal{L}_{\text{con}}
\big(
\mathrm{GE}(G_1),
\mathrm{GE}(G_2),
\mathcal{E}_{\text{LLM}};\Theta
\big),
\end{equation}
where $G_1$ and $G_2$ denote two augmented graph views, $\mathcal{E}_{\text{LLM}}$ represents the token embedding space of the LLM, and $\mathcal{L}_{\text{con}}(\cdot)$ denotes a unified contrastive objective that enforces both cross-modal graph consistency and semantic alignment. The detailed formulation of the loss is provided in Appendix~\ref{Details of Contrastive Pre-training}.

\begin{figure*}[t]
  \centering
  \includegraphics[width=1\columnwidth]{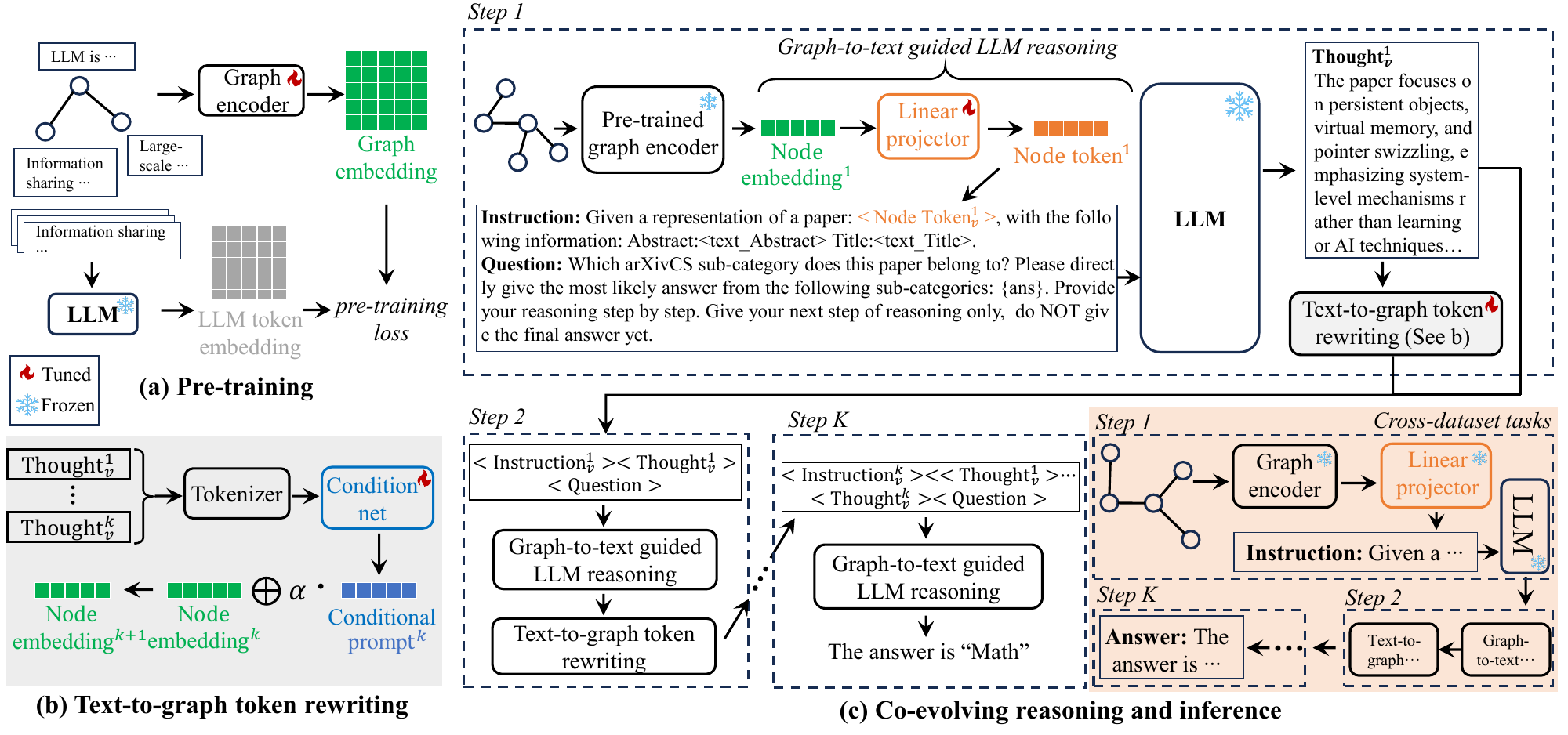}
  \caption{Overall framework of \model.}
  \label{fig.model}
\end{figure*}

\stitle{Graph tokenization.}
To maintain a stable interface to the LLM, a lightweight linear projector is used to map graph embeddings $\vec{H}$ to the LLM embedding space:
\begin{equation}
\vec{H}_{\text{token}} = \mathrm{Linear}(\vec{H};\phi),
\label{eq:graph_token}
\end{equation}
where $\phi$ is the learnable parameter. $\vec{H}_{\text{token}} \in \mathbb{R}^{|V| \times L}$ represents graph token with each row corresponding to a node token of dimension $L$, matching the LLM token dimension.
The node tokens serve as soft prompts that encode graph-based structural information, which are compact, differentiable vectors of graph evidence that can be directly consumed by the LLM to guide step-wise reasoning.

\stitle{Problem definition.}
We study graph transfer learning under a cross-dataset setting, where a model trained on a labeled source graph is directly applied to unseen target graphs without further adaptation.
We consider two tasks. For zero-shot node classification, given a graph $G=(V,E)$ with label space $Y$, the goal is to predict the label $y_i \in Y$ for each node $v_i \in V$ without using any target labels. 
For link prediction, the model adapted on node classification is directly transferred to link prediction without additional task-specific training. Given a node pair $(v_i, v_j)$, the goal is to predict whether an edge exists between them, represented by a binary variable $a_{ij} \in \{0,1\}$. In both settings, the model is trained only on the source dataset and remains frozen during inference on target datasets.
\section{Co-Evolve Chain-of-Thought Prompting}
In this section, we introduce \model, starting with an overview of its framework. Then we detail its core components and conclude with an algorithm and complexity analysis in Appendix~\ref{Algorithm and Complexity Analysis}.

\subsection{Overall Framework}
Fig.~\ref{fig.model} provides an overview of \model, which consists of two phases: graph-text pre-training, and co-evolving CoT reasoning.
We first pre-train a graph encoder following Sect.~\ref{Preliminaries}, as illustrated in Fig.~\ref{fig.model}(a).
Building on the pre-trained encoder and an LLM, \model\ then performs co-evolving CoT prompting to carry out multi-step reasoning before producing the final prediction (Fig.~\ref{fig.model}(c)).
Specifically, each reasoning step alternates between two tightly coupled operations: \emph{graph-to-text guided LLM reasoning} and \emph{text-to-graph token rewriting}.
First, we combine the current graph-token state with lightweight task instructions and the accumulated thoughts, and prompt the LLM to generate an intermediate thought in a step-wise manner. Then the intermediate thought is passed through a condition network to produce node-specific conditioning signals, which are then used to update the graph-token state, as shown in Fig.~\ref{fig.model}(b). Iterating this two-operation cycle enables the structural evidence carried by graph tokens and the LLM’s reasoning trajectory to co-evolve across steps, ultimately yielding the final answer for the downstream task.

\subsection{Graph-to-Text Guided LLM Reasoning}

To guide the LLM toward step-wise reasoning grounded in graph evidence, we design an instruction template with three components: graph token injection, task description, and step-wise prompting. We example with node classification on \textit{arXiv}, as illustrated in Fig.~\ref{fig.model}(c) step 1.

First, for each node, we inject its node token into the instruction together with the node’s textual attributes as follows: \texttt{Given a representation of a paper: <Node token$_v^1$>, with the following information: \textbackslash n Abstract:<text\_Abstract> \textbackslash n Title:<text\_Title>}, where \texttt{<Node token$_v^1$>} is the placeholder for graph inputs, \texttt{<text\_Abstract>} and \texttt{<text\_Title>} contain the node textual content. We define this prompt as \texttt{<Instruction$_v^1$>}.

Second, we formulate the task as selecting the most appropriate label from a candidate set. Using the same node classification task on \emph{arXiv} as an example, the instruction is extended to include the following: \texttt{Which arXiv CS sub-category does this paper belong to? Please give the most likely answer from the following sub-categories: <Ans>}. \texttt{<Ans>} represents the set of alternative answers, which varies across datasets. We define this prompt as \texttt{<Question>}.


Finally, we adopt an iterative prompting strategy to enable multi-step reasoning. At each reasoning step, the prompt is dynamically constructed by augmenting the original instruction (including graph token injection and task description) with previously generated thoughts.
Concretely, at step $k$, the full prompt is constructed as:
\texttt{<Instruction>$_v^k$ <Thought$_v$$^1$> $\cdots$ <Thought$_v$$^{k-1}$> <Question> Provide your reasoning step by step. Give only the next reasoning step, do NOT output the final answer.}
Based on this prompt, the LLM generates the next intermediate thought \texttt{<Thought$_v$$^k$>}. The newly generated thought is then appended to the prompt for subsequent steps, while the graph-token representations are simultaneously updated via the text-to-graph token rewriting module. In this way, both the prompt and the underlying graph representations evolve jointly across reasoning steps.
After completing $K$ reasoning steps, the final prompt is constructed as:
\texttt{<Instruction>$_v^K$ <Thought$_v$$^1$> $\cdots$ <Thought$_v$$^K$> <Question>.}

\subsection{Text-to-Graph Token Rewriting}
\label{Text-to-Graph Token Rewriting}
To tightly couple language reasoning with graph representations, we leverage intermediate thoughts as conditioning signals to dynamically modulate graph tokens in a node-wise manner. 
At reasoning step $k$, the LLM produces an intermediate thought. We then obtain its representation in the LLM embedding space and use it as the reasoning state for subsequent graph rewriting. 
For each node $v$, we collect the representations of thoughts associated with previous reasoning steps, denoted as $\{\vec{t}_v^1,\cdots,\vec{t}_v^k\}$. These representations are fed into a lightweight conditional network to generate a node-specific conditional prompt:
\begin{equation}
    \vec{p}_v^k = \mathrm{CondNet}(\vec{t}_v^1 \oplus \cdots \oplus \vec{t}_v^k; \psi),
\end{equation}
where $\oplus$ denotes concatenation, $\psi$ is the learnable parameter. $\mathrm{CondNet}$ is a lightweight hypernetwork \citep{ha2016hypernetworks,yu2025node} that maps reasoning states into the graph embedding space to generate node-specific conditional prompts. 
In our implementation, $\mathrm{CondNet}$ is instantiated as a two-layer MLP.
For each node $v$, the generated conditional prompt $\vec{p}_v^k$ is used to update its graph embedding through a residual rewriting operation:
\begin{equation}
    \vec{h}_v^{k+1}
    =
    \vec{h}_v^k
    +
    \alpha \cdot \vec{p}_v^k,
    \label{equ:alpha}
\end{equation}
where $\alpha$ is a hyperparameter that controls the update strength. In the next reasoning step, the updated node embedding is then projected again into node token and injected into the instruction. 
This design establishes a closed feedback loop in which intermediate reasoning states continuously rewrite structural representations, enabling progressive refinement of graph evidence across reasoning steps.

\subsection{Adaptation and Cross-dataset Inference}
We adapt \model\ on a labeled dataset while keeping both the graph encoder and the LLM frozen. During this stage, only the linear projector and the condition network are optimized.
Formally, given a labeled dataset $\mathcal{D}=\{(v_i, y_i)\}$, we construct an instruction for each node and prompt the LLM to produce an answer from a candidate label set. 
The adaptation objective is defined as:
\begin{equation}
\mathcal{L}_\text{down}(\phi,\psi) =
-\textstyle\sum_{(v_i,y_i)\in\mathcal{D}}
\log p_{\text{LLM}}(y_i \mid \texttt{Instruction}_{v_i};\phi,\psi),
\end{equation}
where $p_{\text{LLM}}$ denotes the probability of generating the correct answer $y_i$ based on the instruction and the co-evolving reasoning process.

For cross-dataset inference, the learned projector and condition network are directly applied to unseen datasets without any further training.
Given a new dataset or task, we construct the corresponding instruction by injecting graph tokens and specifying the task description. The frozen LLM then performs multi-step reasoning based on this instruction, and outputs the final answer as the prediction.
\section{Experiments}
In this section, we evaluate \model\ and present a comprehensive empirical analysis.

\subsection{Experimental Setup}

\stitle{Datasets.} 
We conduct experiments on eight widely used benchmarks spanning two domains: citation networks and e-commerce graphs. The citation domain includes Arxiv~\citep{hu2020open}, Pubmed~\citep{he2023harnessing}, and an expanded version of Cora~\citep{wen2023augmenting}, where nodes represent papers and edges denote citation relations. The e-commerce domain consists of Children, History, Computer, Photo, and Sports ~\citep{yan2023comprehensive}, where nodes correspond to products and edges capture user behavior correlations.
This diverse selection enables evaluation under varying graph structures, semantics, and domain shifts. Dataset statistics are summarized in Table~\ref{tab:dataset_stats_domain}, with additional dataset details provided in Appendix~\ref{Further Descriptions of Datasets}.

\stitle{Baselines.} 
We compare \model\ against a comprehensive set of baselines spanning multiple paradigms, covering both graph learning and LLM-based reasoning:
(1) \emph{Non-graph model}: MLP~\citep{taud2017multilayer} operates solely on node features and ignores graph structure, serving as a lower-bound baseline.
(2) \emph{Supervised GNNs}: GCN~\citep{kipf2016semi}, GraphSAGE~\citep{hamilton2017inductive}, and GAT~\citep{velivckovic2017graph} learn node representations via neighborhood aggregation under supervised training.
(3) \emph{Self-supervised graph methods}: DGI~\citep{velivckovic2018deep} performs contrastive learning between node representations and global summaries to capture structural information without labels.
(4) \emph{Graph knowledge distillation frameworks}: GKD~\citep{yang2022geometric} and GLNN~\citep{eliasof2024global} transfer knowledge from complex GNNs to simpler models via teacher–student distillation.
(5) \emph{Graph transformers}: NodeFormer~\citep{wu2022nodeformer} and DIFFormer~\citep{wu2023difformer} replace message passing with global attention mechanisms to capture long-range dependencies.
(6) \emph{Graph-only reasoning models}: GCoT~\citep{yu2025gcot} introduces chain-of-thought–style reasoning within graph neural networks by iteratively refining node representations in the latent space.
(7) \emph{LLM-based graph methods}: Vicuna-7B-v1.5~\citep{team2023vicuna} and OFA~\citep{liu2023one} provide general-purpose language or multimodal reasoning backbones, while GraphGPT~\citep{tang2024graphgpt}, LLaGA~\citep{chen2024llaga}, TEA-GLM~\citep{wang2024llms}, and GOFA~\citep{kong2024gofa} integrate graph structures into LLMs via graph tokenization, prompting, or alignment mechanisms. 
Detailed descriptions of baselines are provided in Appendix~\ref{Further Descriptions of Baselines}.

\stitle{Evaluation setting.}
Following the data split and evaluation protocol of \method{TEA-GLM}, we use Arxiv and Computer separately as source datasets for pre-training and downstream adaptation, and evaluate all methods directly on unseen target datasets without additional tuning. 
Specifically, PubMed and Cora are adopted as target datasets in the citation domain, while Children, History, Photo, and Sports are used in the e-commerce domain. 
For the citation-domain transfer setting, the model is trained on 90,941 nodes from Arxiv; for the e-commerce transfer setting, the model is trained on 62,748 nodes from Computer. 
For baseline methods evaluated under the same experimental setting, we directly report the results from \method{TEA-GLM}~\citep{wang2024llms} if available. For \method{GOFA}, whose results are not reported in \method{TEA-GLM}, we reproduce the experiments using the same data splits and evaluation metrics, and report the best-performing results after hyperparameter tuning. In our experiment, we adapt the released pre-trained graph encoder checkpoint provided by \method{TEA-GLM}. Implementation details of \method{GOFA} and \model\ are provided in Appendix~\ref{Implementation Details}.

\subsection{Cross-dataset Performance Evaluation}
We evaluate the cross-dataset generalization ability of \model\ under both node classification and link prediction settings. 
Specifically, models are trained on the source datasets \textit{Arxiv} and \textit{Computer} via node classification task, and directly transferred to unseen target datasets without further adaptation. 

\begin{table}[t]
  \caption{Accuracy of node classification.}
  \label{tab:llm_for_graph_predictor}
  \centering
  \small
  \setlength{\tabcolsep}{2.5pt}
  \renewcommand{\arraystretch}{1.2}
  \resizebox{\columnwidth}{!}{
  \begin{tabular}{llcccccc}
    \toprule
    \multirow{2}{*}{Model type} & \multirow{2}{*}{Model} &
    \multicolumn{2}{c}{Citation} & \multicolumn{4}{c}{E-commerce} \\
    \cmidrule(lr){3-4}\cmidrule(lr){5-8}
    & & Pubmed & Cora & Children & History & Photo & Sports \\
    \midrule

    & \method{MLP} & 0.323$\pm$0.027 & 0.021$\pm$0.006 & 0.029$\pm$0.037 & 0.080$\pm$0.041 & 0.110$\pm$0.070 & 0.042$\pm$0.021 \\
    \midrule

    \multirow{10}{*}{GNN as predictor}
    & \method{GCN}        & 0.288$\pm$0.092 & 0.017$\pm$0.004 & 0.030$\pm$0.018 & 0.063$\pm$0.042 & 0.103$\pm$0.047 & 0.042$\pm$0.025 \\
    & \method{GraphSAGE}  & 0.316$\pm$0.058 & 0.014$\pm$0.007 & 0.008$\pm$0.007 & 0.195$\pm$0.206 & 0.056$\pm$0.055 & 0.051$\pm$0.015 \\
    & \method{GAT}        & 0.343$\pm$0.064 & 0.016$\pm$0.004 & 0.086$\pm$0.084 & 0.172$\pm$0.098 & 0.050$\pm$0.027 & 0.142$\pm$0.138 \\
    & \method{DGI}        & 0.329$\pm$0.103 & 0.026$\pm$0.009 & 0.082$\pm$0.035 & 0.218$\pm$0.168 & 0.224$\pm$0.127 & 0.049$\pm$0.017 \\
    & \method{GKD}        & 0.399$\pm$0.033 & 0.042$\pm$0.008 & 0.202$\pm$0.064 & 0.339$\pm$0.138 & 0.166$\pm$0.086 & 0.208$\pm$0.077 \\
    & \method{GLNN}       & 0.390$\pm$0.011 & 0.031$\pm$0.006 & 0.187$\pm$0.012 & 0.283$\pm$0.021 & 0.403$\pm$0.019 & 0.317$\pm$0.048 \\
    & \method{NodeFormer} & 0.308$\pm$0.093 & 0.016$\pm$0.007 & 0.048$\pm$0.028 & 0.168$\pm$0.127 & 0.073$\pm$0.015 & 0.165$\pm$0.057 \\
    & \method{DIFFormer}  & 0.361$\pm$0.071 & 0.029$\pm$0.014 & 0.129$\pm$0.030 & 0.275$\pm$0.171 & 0.321$\pm$0.055 & 0.306$\pm$0.131 \\
    & \method{OFA}        & 0.314$\pm$0.059 & 0.130$\pm$0.019 & 0.064$\pm$0.086 & 0.052$\pm$0.049 & 0.340$\pm$0.026 & 0.101$\pm$0.071 \\
    & \method{GCoT}       & 0.378$\pm$0.025 & 0.072$\pm$0.056 & 0.159$\pm$0.042 & 0.374$\pm$0.048 & 0.373$\pm$0.037 & 0.257$\pm$0.051 \\
    \midrule

    \multirow{8}{*}{LLM as predictor}
    & \method{Vicuna-7B-v1.5} & 0.719$\pm$0.010 & 0.156$\pm$0.001 & 0.270$\pm$0.001 & 0.363$\pm$0.001 & 0.378$\pm$0.004 & 0.370$\pm$0.001 \\
    & \method{Vicuna-7B-SPT}  & 0.768$\pm$0.036 & 0.168$\pm$0.018 & 0.227$\pm$0.015 & 0.281$\pm$0.088 & 0.350$\pm$0.061 & 0.230$\pm$0.018 \\
    & \method{GraphGPT-std}   & 0.701 & 0.126 & -- & -- & -- & -- \\
    & \method{GraphGPT-cot}   & 0.521 & 0.181 & -- & -- & -- & -- \\
    & \method{LLaGA}          & 0.793$\pm$0.036 & 0.168$\pm$0.032 & 0.199$\pm$0.007 & 0.146$\pm$0.067 & 0.276$\pm$0.069 & 0.352$\pm$0.033 \\
    & \method{TEA-GLM}        & 0.848$\pm$0.010 & \underline{0.202$\pm$0.014} & \underline{0.271$\pm$0.010} & 0.528$\pm$0.058 & 0.497$\pm$0.027 & 0.404$\pm$0.010 \\
    & \method{GOFA}           & \underline{0.859$\pm$0.028} & 0.172$\pm$0.005 & 0.257$\pm$0.021 & \underline{0.558$\pm$0.028} & \underline{0.512$\pm$0.016} & \underline{0.425$\pm$0.031} \\
    & \model                  & \textbf{0.902$\pm$0.035} & \textbf{0.265$\pm$0.014} & \textbf{0.295$\pm$0.003} & \textbf{0.641$\pm$0.056} & \textbf{0.538$\pm$0.052} & \textbf{0.458$\pm$0.073} \\
    \bottomrule
  \end{tabular}}
  \parbox{\linewidth}{\footnotesize \ \ The best method is bolded and the runner-up is underlined.}
\end{table}

\stitle{Node classification performance.}
Table~\ref{tab:llm_for_graph_predictor} shows that \model\ consistently achieves the best performance across all datasets, demonstrating strong cross-dataset generalization. 
First, traditional GNNs exhibit limited transferability, with performance dropping significantly under distribution shift. This is because their representations are tightly coupled to training graphs through localized message passing, making them sensitive to structural and semantic discrepancies across datasets. This limitation is particularly evident in e-commerce datasets, where node semantics are weakly aligned and relational patterns are more heterogeneous.
Second, pure LLM-based methods demonstrate stronger robustness due to their semantic generalization ability. However, their lack of explicit structural modeling leads to suboptimal performance.
Third, compared with recent graph--LLM approaches such as \method{TEA-GLM} and \method{GOFA}, \model\ achieves consistent improvements across all datasets. The gains are more pronounced on e-commerce datasets, which involve larger domain shifts. This suggests that merely injecting graph information into LLMs is insufficient; instead, effective bidirectional interaction between reasoning and structure is crucial for robust generalization.
Additional Macro-F1 results are provided in the Appendix~\ref{F1 score on node classification task}.
These results indicate that \model\ better balances semantic reasoning and structural modeling, leading to improved transferability under cross-dataset settings.

\begin{wraptable}{r}{0.6\columnwidth}
\vspace{-1.0em}
\centering
\caption{AUC of link prediction.}
\label{tab:llm_graph_generalization}
\small
\setlength{\tabcolsep}{2.5pt}
\renewcommand{\arraystretch}{0.9}
\resizebox{\linewidth}{!}{
\begin{tabular}{lccc@{\hspace{6pt}}ccccc}
\toprule
\multirow{2}{*}{Model} &
\multicolumn{3}{c}{Citation} &
\multicolumn{5}{c}{E-commerce} \\
\cmidrule(lr){2-4} \cmidrule(lr){5-9}
& Arxiv & Pubmed & Cora
& Children & History & Computer & Photo & Sports \\
\midrule
\method{OFA}            & 0.469 & 0.481 & 0.492 & 0.484 & 0.431 & 0.461 & 0.459 & 0.517 \\
\method{Vicuna-7B-v1.5} & 0.513 & 0.543 & 0.527 & 0.500 & 0.515 & 0.502 & 0.501 & 0.502 \\
\method{Vicuna-7B-SPT}  & 0.537 & 0.535 & 0.565 & 0.544 & 0.543 & 0.509 & 0.501 & 0.508 \\
\method{GraphGPT-std}   & 0.649 & 0.501 & 0.520 & --    & --    & --    & --    & --    \\
\method{LLaGA}          & 0.570 & 0.569 & 0.537 & 0.422 & 0.449 & 0.479 & 0.478 & \underline{0.597} \\
\method{TEA-GLM}        & \underline{0.657} & \underline{0.689} & \underline{0.586} & \underline{0.571} & \underline{0.579} & \underline{0.554} & \underline{0.545} & 0.553 \\
\method{GOFA}           & 0.635 & 0.627 & 0.536 & 0.547 & 0.561 & 0.549 & 0.538 & 0.571 \\
\midrule
\textbf{\model}         & \textbf{0.724} & \textbf{0.762} & \textbf{0.645} & \textbf{0.624} & \textbf{0.656} & \textbf{0.642} & \textbf{0.549} & \textbf{0.629} \\
\bottomrule
\end{tabular}}
\vspace{-1.0em}
\end{wraptable}

\stitle{Link prediction performance.}
Table~\ref{tab:llm_graph_generalization} reports the AUC results for cross-task link prediction, where models trained on node classification are directly evaluated on link prediction without any task-specific adaptation. This setting provides a stringent test of whether learned representations are task-agnostic.
\model\ consistently outperforms all baselines across both domains, indicating strong cross-task transferability. 
Pure LLM-based methods achieve moderate performance due to their semantic reasoning capability, but their inability to explicitly model graph structure limits their effectiveness in edge-level prediction tasks. Existing hybrid approaches improve upon this by incorporating structural signals, yet they largely treat graph representations as static inputs.
In contrast, \model\ enables dynamic interaction between reasoning and graph representations, allowing node-level semantic signals to iteratively reshape structural embeddings. This leads to more coherent representations that generalize beyond node classification and transfer effectively to link prediction.

\subsection{Ablation Study}
To further verify the effectiveness of the proposed co-evolving design, we compare \model\ with two ablation variants in Table~\ref{tab:ablation_combined}. Variant 1 corresponds to a single-step reasoning model, which is equivalent to \method{TEA-GLM} and does not perform multi-step reasoning. Variant 2 introduces multi-step CoT reasoning through Graph-to-Text Guided LLM Reasoning, but removes the Text-to-Graph Token Rewriting module, such that graph embeddings remain fixed throughout inference. As shown in Table~\ref{tab:ablation_combined}, \model\ consistently achieves the best performance on cross-dataset node classification across all target datasets, with especially clear improvements on datasets such as Pubmed and History, indicating stronger robustness under structural and semantic distribution shifts. A similar trend is observed for cross-task link prediction, where \model\ also outperforms both variants on all datasets when directly transferred from node classification without any task-specific adaptation. These results suggest that multi-step reasoning alone is beneficial, but insufficient to fully unleash the transferability of the model when graph evidence remains static. In contrast, explicitly feeding intermediate reasoning signals back into graph representations allows the model to iteratively refine structural evidence, leading to more transferable graph embeddings. Overall, the ablation results confirm that the advantage of \model\ comes from the complete co-evolving reasoning process rather than any single component alone.

\begin{table*}[h]
\centering
\caption{Ablation study results on node classification and link prediction.}
\label{tab:ablation_combined}
\scriptsize
\setlength{\tabcolsep}{2pt}
\resizebox{\textwidth}{!}{
\begin{tabular}{lcccccc@{\hspace{10pt}}cccccccc}
\toprule
\multirow{3}{*}{Model} &
\multicolumn{6}{c}{Node classification (accuracy)} &
\multicolumn{8}{c}{Link prediction (AUC)} \\
\cmidrule(lr){2-7} \cmidrule(lr){8-15}
&
\multicolumn{2}{c}{Citation} &
\multicolumn{4}{c}{E-commerce} &
\multicolumn{3}{c}{Citation} &
\multicolumn{5}{c}{E-commerce} \\
\cmidrule(lr){2-3} \cmidrule(lr){4-7}
\cmidrule(lr){8-10} \cmidrule(lr){11-15}
&
Pubmed & Cora & Children & History & Photo & Sports
& Arxiv & Pubmed & Cora & Children & History & Computer & Photo & Sports \\
\midrule
\method{Variant 1}
& 0.848 & 0.202 & \underline{0.271} & 0.528 & \underline{0.497} & 0.404
& 0.657 & 0.689 & 0.586 & 0.571 & 0.579 & 0.554 & 0.545 & 0.553 \\

\method{Variant 2}
& \underline{0.861} & \underline{0.231} & 0.251 & \underline{0.551} & 0.493 & \underline{0.435}
& \underline{0.683} & \underline{0.702} & \underline{0.591} & \underline{0.598} & \underline{0.621} & \underline{0.614} & \underline{0.547} & \underline{0.572} \\
\textbf{\model}
& \textbf{0.902} & \textbf{0.265} & \textbf{0.295} & \textbf{0.641} & \textbf{0.538} & \textbf{0.458}
& \textbf{0.724} & \textbf{0.762} & \textbf{0.645} & \textbf{0.624} & \textbf{0.656} & \textbf{0.642} & \textbf{0.549} & \textbf{0.629} \\
\bottomrule
\end{tabular}}
\end{table*}




\subsection{Hyperparameter Analysis}

We analyze key hyperparameters in \model\ to justify our design choices.


\begin{figure}[h]
    \centering

    \begin{minipage}[t]{0.49\columnwidth}
        \centering
        \includegraphics[width=\linewidth]{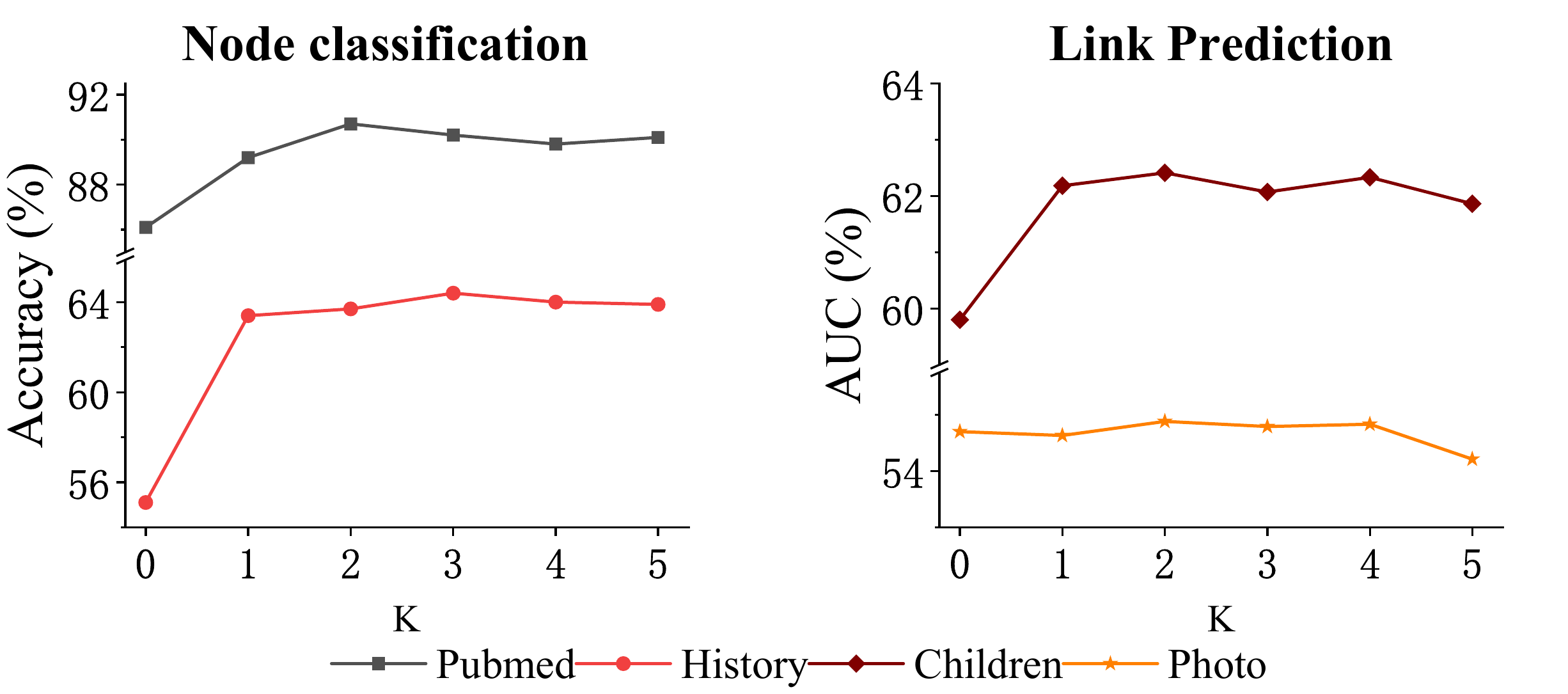}
        \caption{Effect of total inference steps $K$.}
        \label{fig:infer_steps}
    \end{minipage}
    \hfill
    \begin{minipage}[t]{0.49\columnwidth}
        \centering
        \includegraphics[width=\linewidth]{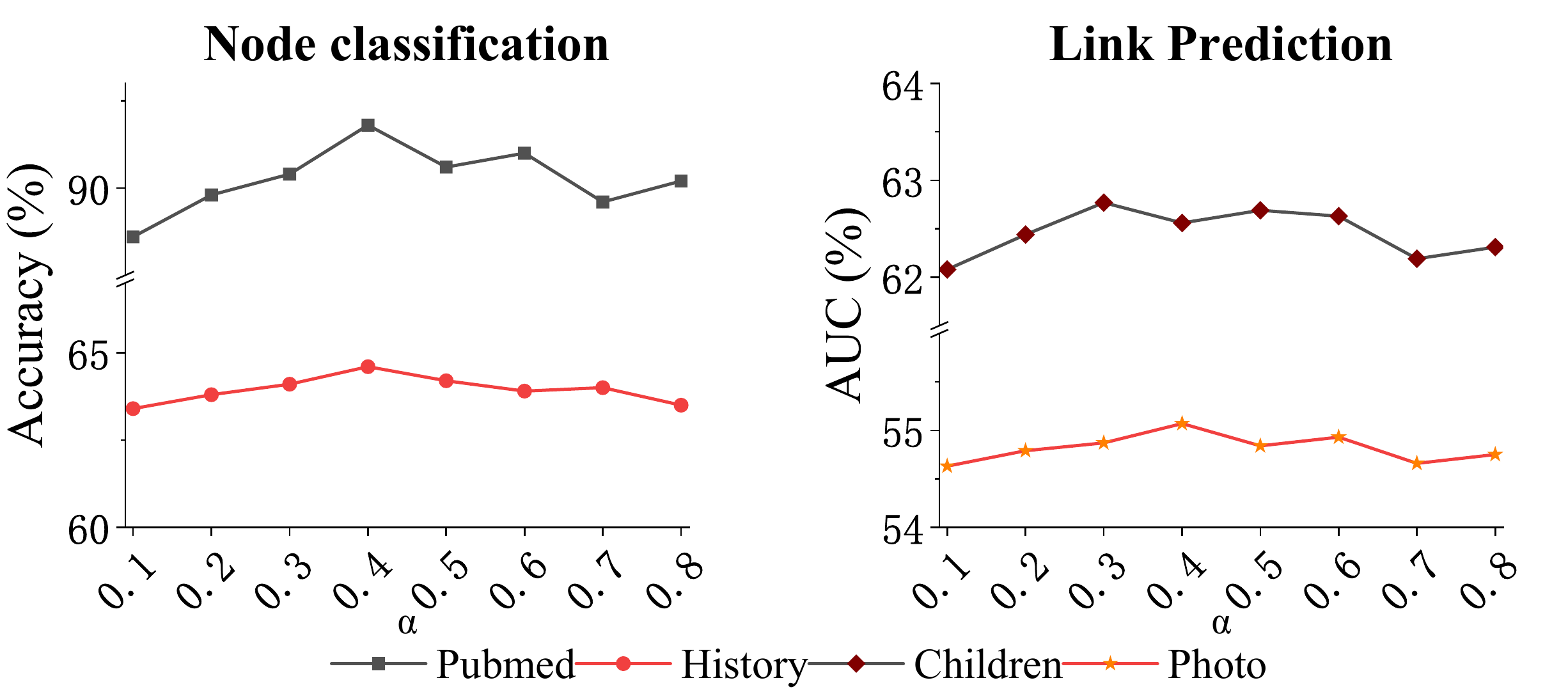}
        \caption{Effect of the condition weight $\alpha$.}
        \label{fig:alpha}
    \end{minipage}

\end{figure}

\stitle{Number of inference steps.}
We further vary the number of inference steps $K$, with results shown in Fig.~\ref{fig:infer_steps}. For node classification, performance improves from $K=1$ to small values (typically $K=2$), but does not benefit from further increasing $K$. This indicates that a small number of reasoning steps is sufficient to capture task-relevant information, while deeper reasoning may introduce unnecessary complexity or noise.
For link prediction, performance remains relatively stable across different values of $K$, suggesting that the task is less sensitive to reasoning depth. This may be because link prediction primarily depends on structural compatibility, which can be adequately captured with limited reasoning iterations. Based on these results, we set $K=2$ as a default choice, achieving a good balance between performance and efficiency.


\stitle{Condition weight of rewriting.}
We further vary the condition weight $\alpha$ in Eq.~\eqref{equ:alpha}, which controls the update strength of thought-conditioned rewriting, and report the results in Fig.~\ref{fig:alpha}. The results show that performance generally improves as $\alpha$ increases from small values, indicating that a moderate update strength is important for effectively injecting reasoning signals into graph representations. When $\alpha$ is too small, the rewriting signal is weak and cannot sufficiently refine the graph tokens. As $\alpha$ becomes larger, the performance no longer improves consistently and instead shows mild fluctuations, suggesting that overly strong updates may disturb the structural evidence and reduce the stability of representation refinement. Overall, the best or near-best performance is consistently achieved around $\alpha=0.4$ across both node classification and link prediction, demonstrating that a moderate rewriting strength provides the best trade-off between effective conditioning and stable representation evolution. Based on this observation, we set $\alpha=0.4$ in all experiments.

\begin{wrapfigure}[10]{r}{0.50\columnwidth}
    \vspace{-1.0em}
    \centering
    \includegraphics[width=\linewidth]{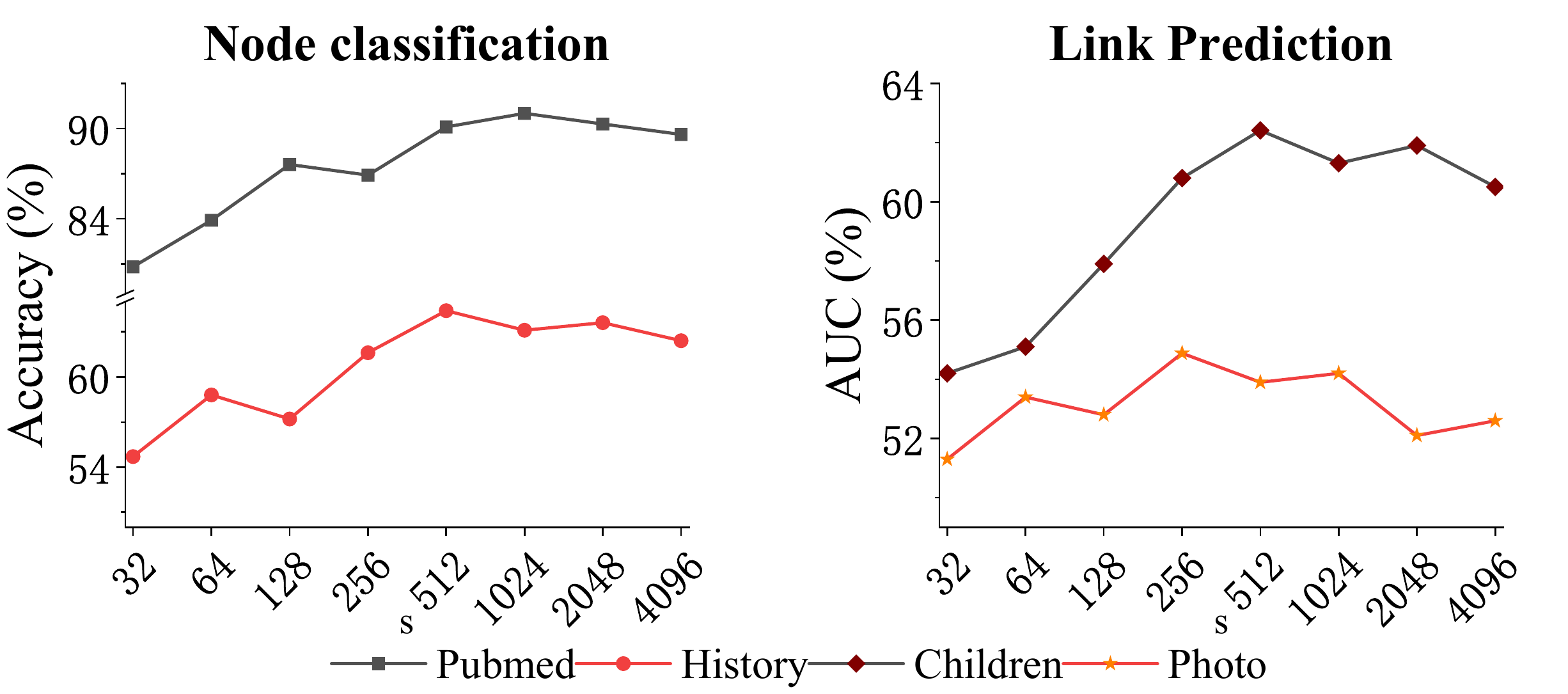}
    \caption{Effect of hidden dimension $s$.}
    \label{fig:hidden_dim}
    \vspace{-1.0em}
\end{wrapfigure}

\stitle{Hidden dimension of condition network.}
We implement the condition network as a two-layer MLP and vary its hidden dimension $s$, with the results reported in Fig.~\ref{fig:hidden_dim}. Here, $s$ controls only the internal transformation capacity of the condition network, while its output dimension is fixed to the graph encoder output dimension. This ensures that the generated conditional update can be added to the graph embeddings through the residual operation in Eq.~\eqref{equ:alpha}. The results show that performance consistently improves as $s$ increases from small values, indicating that a higher-dimensional space enhances the model’s ability to encode reasoning signals for effective text-to-graph rewriting. 
However, the performance saturates when $s$ reaches a moderate range (e.g., $s=512$ or $1024$), and slightly degrades for larger values. This suggests that while insufficient capacity limits the expressiveness of the conditioning signals, overly large dimensions introduce redundant parameters and may lead to noisy or less stable updates to graph embeddings. Based on this observation, we adopt $s=512$ in all experiments.

    \label{fig:cot_vis}



\section{Visualization}
To further analyze how \model\ refines graph representations through reasoning, we visualize the evolution of node embeddings on Arxiv, with the corresponding results on History provided in the Appendix~\ref{Visualization}. As shown in Fig.~\ref{fig:cot_vis}, the initial embeddings are highly entangled across classes, indicating limited class separability under distribution shift. After the first reasoning step, the embeddings begin to form coarse clusters, suggesting that intermediate thoughts introduce task-relevant semantic signals. After the second reasoning step, the clusters become more compact and better separated, showing that Text-to-Graph Token Rewriting progressively reshapes the embedding space. This step-wise evolution demonstrates that reasoning signals can iteratively refine graph representations, leading to more discriminative embeddings for downstream prediction.

\begin{figure}[h]
    \centering
    \includegraphics[width=0.7\textwidth]{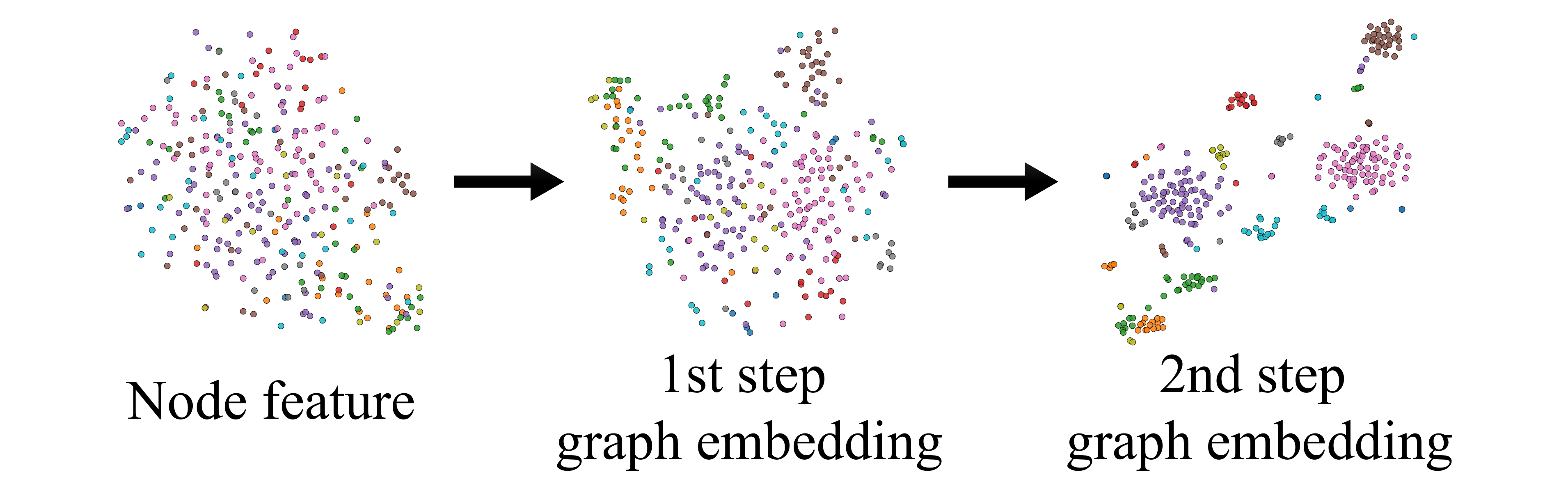}
    \caption{Visualization of step-wise node representation evolution on ArXiv.}
    \label{fig:cot_vis}
\end{figure}

\section{Conclusion}
In this paper, we proposed \model, a CoT prompting framework that enables language reasoning and structural evidence to co-evolve in a closed feedback loop. We designed a text-to-graph token rewriting mechanism that transforms intermediate thoughts into actionable controls to rewrite graph tokens step by step, and then fed the updated tokens back into subsequent instructions, achieving graph-to-text guided LLM reasoning. Extensive experiments demonstrated that \model\ consistently outperforms strong state-of-the-art baselines across challenging cross-dataset transfer setting.

\newpage
\bibliographystyle{unsrt}
\bibliography{sample-base}

@inproceedings{feng2023extending,
  title={Extending the design space of graph neural networks by rethinking folklore Weisfeiler-Lehman},
  author={Feng, Jiarui and Kong, Lecheng and Liu, Hao and Tao, Dacheng and Li, Fuhai and Zhang, Muhan and Chen, Yixin},
  booktitle={NeurIPS},
  volume={36},
  year={2023}
}

@inproceedings{kong2023mag,
  title={Mag-gnn: Reinforcement learning boosted graph neural network},
  author={Kong, Lecheng and Feng, Jiarui and Liu, Hao and Tao, Dacheng and Chen, Yixin and Zhang, Muhan},
  booktitle={NeurIPS},
  volume={36},
  pages={12000--12021},
  year={2023}
}

@inproceedings{wang2024gnnboundary,
title={{GNNB}oundary: Towards Explaining Graph Neural Networks through the Lens of Decision Boundaries},
author={Xiaoqi Wang and Han Wei Shen},
booktitle={ICLR},
year={2024}
}

@inproceedings{pahng2024improving,
  title={Improving graph neural networks by learning continuous edge directions},
  author={Pahng, Seong Ho and Hormoz, Sahand},
  booktitle={ICLR},
  year={2024}
}

@inproceedings{zhao2025graph,
  title={Graph Learning with Distributional Edge Layouts},
  author={Zhao, Xinjian and Ying, Chaolong and Xu, Yaoyao and Yu, Tianshu},
  booktitle={SIGKDD},
  pages={2055--2066},
  year={2025}
}

@inproceedings{bechler2309graph,
  title={Graph neural networks use graphs when they shouldn’t},
  author={Bechler-Speicher, Maya and Amos, Ido and Gilad-Bachrach, Ran and Globerson, Amir},
  booktitle={ICML},
  year={2024}
}

@inproceedings{feng2022powerful,
  title={How powerful are k-hop message passing graph neural networks},
  author={Feng, Jiarui and Chen, Yixin and Li, Fuhai and Sarkar, Anindya and Zhang, Muhan},
  booktitle={NeurIPS},
  volume={35},
  year={2022}
}

@inproceedings{xu2018powerful,
  title={How powerful are graph neural networks?},
  author={Xu, Keyulu and Hu, Weihua and Leskovec, Jure and Jegelka, Stefanie},
  booktitle={ICLR},
  year={2018}
}

@inproceedings{alon2020bottleneck,
  title={On the bottleneck of graph neural networks and its practical implications},
  author={Alon, Uri and Yahav, Eran},
  booktitle={ICLR},
  year={2021}
}

@inproceedings{velivckovic2018deep,
  title={Deep graph infomax},
  author={Veli{\v{c}}kovi{\'c}, Petar and Fedus, William and Hamilton, William L and Li{\`o}, Pietro and Bengio, Yoshua and Hjelm, R Devon},
  booktitle={ICLR},
  year={2019}
}

@inproceedings{qiu2020gcc,
  title={Gcc: Graph contrastive coding for graph neural network pre-training},
  author={Qiu, Jiezhong and Chen, Qibin and Dong, Yuxiao and Zhang, Jing and Yang, Hongxia and Ding, Ming and Wang, Kuansan and Tang, Jie},
  booktitle={SIGKDD},
  pages={1150--1160},
  year={2020}
}

@inproceedings{you2021graph,
  title={Graph contrastive learning automated},
  author={You, Yuning and Chen, Tianlong and Shen, Yang and Wang, Zhangyang},
  booktitle={ICML},
  pages={12121--12132},
  year={2021}
}

@inproceedings{hou2022graphmae,
  title={Graphmae: Self-supervised masked graph autoencoders},
  author={Hou, Zhenyu and Liu, Xiao and Cen, Yukuo and Dong, Yuxiao and Yang, Hongxia and Wang, Chunjie and Tang, Jie},
  booktitle={SIGKDD},
  pages={594--604},
  year={2022}
}

@inproceedings{he2025unigraph,
  title={Unigraph: Learning a unified cross-domain foundation model for text-attributed graphs},
  author={He, Yufei and Sui, Yuan and He, Xiaoxin and Hooi, Bryan},
  booktitle={SIGKDD},
  pages={448--459},
  year={2025}
}

@inproceedings{tian2024graph,
  title={Graph neural prompting with large language models},
  author={Tian, Yijun and Song, Huan and Wang, Zichen and Wang, Haozhu and Hu, Ziqing and Wang, Fang and Chawla, Nitesh V and Xu, Panpan},
  booktitle={AAAI},
  volume={38},
  number={17},
  pages={19080--19088},
  year={2024}
}

@inproceedings{hu2025large,
  title={Large language model meets graph neural network in knowledge distillation},
  author={Hu, Shengxiang and Zou, Guobing and Yang, Song and Lin, Shiyi and Gan, Yanglan and Zhang, Bofeng and Chen, Yixin},
  booktitle={AAAI},
  volume={39},
  pages={17295--17304},
  year={2025}
}

@inproceedings{yu2025node,
  title={Node-Time Conditional Prompt Learning In Dynamic Graphs},
  author={Yu, Xingtong and Liu, Zhenghao and Zhang, Xinming and Fang, Yuan},
  booktitle={ICLR},
  year={2025}
}

@inproceedings{zolnai2024stage,
  title={Stage: Simplified text-attributed graph embeddings using pre-trained llms},
  author={Zolnai-Lucas, Aaron and Boylan, Jack and Hokamp, Chris and Ghaffari, Parsa},
  booktitle={KaLLM},
  year={2024}
}

@inproceedings{qiao2025login,
  title={Login: A large language model consulted graph neural network training framework},
  author={Qiao, Yiran and Ao, Xiang and Liu, Yang and Xu, Jiarong and Sun, Xiaoqian and He, Qing},
  booktitle={WSDM},
  pages={232--241},
  year={2025}
}

@article{sun2023graph,
  title={Graph prompt learning: A comprehensive survey and beyond},
  author={Sun, Xiangguo and Zhang, Jiawen and Wu, Xixi and Cheng, Hong and Xiong, Yun and Li, Jia},
  journal={arXiv preprint arXiv:2311.16534},
  year={2023}
}

@inproceedings{sun2022gppt,
  title={Gppt: Graph pre-training and prompt tuning to generalize graph neural networks},
  author={Sun, Mingchen and Zhou, Kaixiong and He, Xin and Wang, Ying and Wang, Xin},
  booktitle={SIGKDD},
  pages={1717--1727},
  year={2022}
}

@inproceedings{fang2023universal,
  title={Universal prompt tuning for graph neural networks},
  author={Fang, Taoran and Zhang, Yunchao and Yang, Yang and Wang, Chunping and Chen, Lei},
  booktitle={NeurIPS},
  volume={36},
  year={2023}
}

@inproceedings{zhao2023graphtext,
  title={Graphtext: Graph reasoning in text space},
  author={Zhao, Jianan and Zhuo, Le and Shen, Yikang and Qu, Meng and Liu, Kai and Bronstein, Michael and Zhu, Zhaocheng and Tang, Jian},
  booktitle={NeurIPS},
  year={2024}
}

@inproceedings{tran2025groot,
  title={GROOT: Effective Design of Biological Sequences with Limited Experimental Data},
  author={Tran, Thanh VT and Ngo, Nhat Khang and Nguyen, Viet Anh and Hy, Truong Son},
  booktitle={SIGKDD},
  pages={1385--1396},
  year={2025}
}

@inproceedings{zhang2024graphtranslator,
  title={Graphtranslator: Aligning graph model to large language model for open-ended tasks},
  author={Zhang, Mengmei and Sun, Mingwei and Wang, Peng and Fan, Shen and Mo, Yanhu and Xu, Xiaoxiao and Liu, Hong and Yang, Cheng and Shi, Chuan},
  booktitle={WWW},
  pages={1003--1014},
  year={2024}
}

@article{yu2025graph2text,
  title={Graph2text or graph2token: A perspective of large language models for graph learning},
  author={Yu, Shuo and Wang, Yingbo and Li, Ruolin and Liu, Guchun and Shen, Yanming and Ji, Shaoxiong and Li, Bowen and Han, Fengling and Zhang, Xiuzhen and Xia, Feng},
  journal={ACM},
  year={2026},
  publisher={ACM New York, NY}
}

@inproceedings{kong2024gofa,
  title={GOFA: A Generative One-For-All Model for Joint Graph Language Modeling},
  author={Kong, Lecheng and Feng, Jiarui and Liu, Hao and Huang, Chengsong and Huang, Jiaxin and Chen, Yixin and Zhang, Muhan},
  booktitle={ICLR},
  year={2025}
}

@inproceedings{kipf2016semi,
  title={Semi-supervised classification with graph convolutional networks},
  author={Kipf, TN},
  booktitle={ICLR},
  year={2017}
}

@inproceedings{hamilton2017inductive,
  title={Inductive representation learning on large graphs},
  author={Hamilton, Will and Ying, Zhitao and Leskovec, Jure},
  booktitle={NeurIPS},
  volume={30},
  year={2017}
}

@inproceedings{velivckovic2017graph,
  title={Graph attention networks},
  author={Veli{\v{c}}kovi{\'c}, Petar and Cucurull, Guillem and Casanova, Arantxa and Romero, Adriana and Lio, Pietro and Bengio, Yoshua},
  booktitle={ICLR},
  year={2018}
}

@inproceedings{yang2022geometric,
  title={Geometric knowledge distillation: Topology compression for graph neural networks},
  author={Yang, Chenxiao and Wu, Qitian and Yan, Junchi},
  booktitle={NeurIPS},
  volume={35},
  pages={29761--29775},
  year={2022}
}

@article{eliasof2024global,
  title={Global-local graph neural networks for node-classification},
  author={Eliasof, Moshe and Treister, Eran},
  journal={Pattern Recognition Letters},
  volume={184},
  pages={103--110},
  year={2024},
  publisher={Elsevier}
}

@inproceedings{wu2022nodeformer,
  title={Nodeformer: A scalable graph structure learning transformer for node classification},
  author={Wu, Qitian and Zhao, Wentao and Li, Zenan and Wipf, David P and Yan, Junchi},
  booktitle={NeurIPS},
  volume={35},
  pages={27387--27401},
  year={2022}
}

@inproceedings{wu2023difformer,
  title={Difformer: Scalable (graph) transformers induced by energy constrained diffusion},
  author={Wu, Qitian and Yang, Chenxiao and Zhao, Wentao and He, Yixuan and Wipf, David and Yan, Junchi},
  booktitle={ICLR},
  year={2023}
}

@inproceedings{liu2023one,
  title={One for all: Towards training one graph model for all classification tasks},
  author={Liu, Hao and Feng, Jiarui and Kong, Lecheng and Liang, Ningyue and Tao, Dacheng and Chen, Yixin and Zhang, Muhan},
  booktitle={ICLR},
  year={2024}
}

@inproceedings{yu2025gcot,
  title={GCoT: Chain-of-thought prompt learning for graphs},
  author={Yu, Xingtong and Zhou, Chang and Kuai, Zhongwei and Zhang, Xinming and Fang, Yuan},
  booktitle={SIGKDD},
  pages={3669--3679},
  year={2025}
}

@misc{team2023vicuna,
    title = {Vicuna: An Open-Source Chatbot Impressing GPT-4 with 90\%* ChatGPT Quality},
    url = {https://lmsys.org/blog/2023-03-30-vicuna/},
    author = {Chiang, Wei-Lin and Li, Zhuohan and Lin, Zi and Sheng, Ying and Wu, Zhanghao and Zhang, Hao and Zheng, Lianmin and Zhuang, Siyuan and Zhuang, Yonghao and Gonzalez, Joseph E. and Stoica, Ion and Xing, Eric P.},
    year = {2023}
}

@inproceedings{tang2024graphgpt,
  title={Graphgpt: Graph instruction tuning for large language models},
  author={Tang, Jiabin and Yang, Yuhao and Wei, Wei and Shi, Lei and Su, Lixin and Cheng, Suqi and Yin, Dawei and Huang, Chao},
  booktitle={SIGIR},
  pages={491--500},
  year={2024}
}

@inproceedings{Jin2024GraphCA,
  title={Graph Chain-of-Thought: Augmenting Large Language Models by Reasoning on Graphs},
  author={Bowen Jin and Chulin Xie and Jiawei Zhang and Kashob Kumar Roy and Yu Zhang and Suhang Wang and Yu Meng and Jiawei Han},
  booktitle={ACL},
  year={2024}
}

@article{scarselli2008graph,
  title={The graph neural network model},
  author={Scarselli, Franco and Gori, Marco and Tsoi, Ah Chung and Hagenbuchner, Markus and Monfardini, Gabriele},
  journal={IEEE transactions on neural networks},
  volume={20},
  number={1},
  pages={61--80},
  year={2008},
  publisher={IEEE}
}

@inproceedings{wei2022chain,
  booktitle={Chain-of-thought prompting elicits reasoning in large language models},
  author={Wei, Jason and Wang, Xuezhi and Schuurmans, Dale and Bosma, Maarten and Xia, Fei and Chi, Ed and Le, Quoc V and Zhou, Denny and others},
  journal={NeurIPS},
  volume={35},
  pages={24824--24837},
  year={2022}
}

@inproceedings{wang2023self,
  booktitle={Self-consistency improves chain of thought reasoning in language models},
  author={Wang, Xuezhi and Wei, Jason and Schuurmans, Dale and Le, Quoc and Chi, Ed and Narang, Sharan and Chowdhery, Aakanksha and Zhou, Denny},
  journal={ICLR},
  year={2023}
}

@inproceedings{chen2024llaga,
  title={LLaGA: Large Language and Graph Assistant},
  author={Chen, Runjin and Zhao, Tong and Jaiswal, Ajay and Shah, Neil and Wang, Zhangyang},
  booktitle={ICML},
  year={2024}
}

@inproceedings{wang2024llms,
  title={Llms as zero-shot graph learners: Alignment of gnn representations with llm token embeddings},
  author={Wang, Duo and Zuo, Yuan and Li, Fengzhi and Wu, Junjie},
  booktitle={NeurIPS},
  volume={37},
  pages={5950--5973},
  year={2024}
}

@inproceedings{hu2020open,
  title={Open graph benchmark: Datasets for machine learning on graphs},
  author={Hu, Weihua and Fey, Matthias and Zitnik, Marinka and Dong, Yuxiao and Ren, Hongyu and Liu, Bowen and Catasta, Michele and Leskovec, Jure},
  booktitle={NeurIPS},
  volume={33},
  pages={22118--22133},
  year={2020}
}

@inproceedings{you2020graph,
  booktitle={Graph contrastive learning with augmentations},
  author={You, Yuning and Chen, Tianlong and Sui, Yongduo and Chen, Ting and Wang, Zhangyang and Shen, Yang},
  journal={NeurIPS},
  volume={33},
  pages={5812--5823},
  year={2020}
}

@inproceedings{velickovic2019deep,
  title={Deep Graph Infomax},
  author={Velickovic, Petar and Fedus, William and Hamilton, William L and Li{\`o}, Pietro and Bengio, Yoshua and Hjelm, R Devon},
  booktitle={ICLR},
  year={2019}
}

@inproceedings{liu2023graphprompt,
  title={{GraphPrompt}: Unifying pre-training and downstream tasks for graph neural networks},
  author={Liu, Zemin and Yu, Xingtong and Fang, Yuan and Zhang, Xinming},
  booktitle={WWW},
  pages={417--428},
  year={2023}
}

@article{yu2024few,
  title={Few-Shot Learning on Graphs: from Meta-learning to Pre-training and Prompting},
  author={Yu, Xingtong and Fang, Yuan and Liu, Zemin and Wu, Yuxia and Wen, Zhihao and Bo, Jianyuan and Zhang, Xinming and Hoi, Steven CH},
  journal={arXiv preprint arXiv:2402.01440},
  year={2024}
}

@article{yu2024generalized,
  title={Generalized graph prompt: Toward a unification of pre-training and downstream tasks on graphs},
  author={Yu, Xingtong and Liu, Zhenghao and Fang, Yuan and Liu, Zemin and Chen, Sihong and Zhang, Xinming},
  journal={IEEE TKDE},
  year={2024}
}

@inproceedings{he2023harnessing,
  title={Harnessing explanations: Llm-to-lm interpreter for enhanced text-attributed graph representation learning},
  author={He, Xiaoxin and Bresson, Xavier and Laurent, Thomas and Perold, Adam and LeCun, Yann and Hooi, Bryan},
  booktitle={ICLR},
  year={2024}
}

@inproceedings{wen2023augmenting,
  title={Augmenting low-resource text classification with graph-grounded pre-training and prompting},
  author={Wen, Zhihao and Fang, Yuan},
  booktitle={SIGIR},
  pages={506--516},
  year={2023}
}

@inproceedings{yan2023comprehensive,
  title={A comprehensive study on text-attributed graphs: Benchmarking and rethinking},
  author={Yan, Hao and Li, Chaozhuo and Long, Ruosong and Yan, Chao and Zhao, Jianan and Zhuang, Wenwen and Yin, Jun and Zhang, Peiyan and Han, Weihao and Sun, Hao and others},
  booktitle={NeurIPS},
  volume={36},
  pages={17238--17264},
  year={2023}
}

@inproceedings{10.1145/3711896.3737031,
author = {Wo, Zengyi and Shao, Minglai and Zhang, Shiyu and Wang, Ruijie},
title = {Local Homophily-Aware Graph Neural Network with Adaptive Polynomial Filters for Scalable Graph Anomaly Detection},
booktitle = {SIGKDD},
pages = {3180–3191},
year = {2025}
}

@inproceedings{10.1145/3690624.3709192,
author = {Higashiguchi, Shingo and Matsubara, Yasuko and Kawabata, Koki and Murayama, Taichi and Sakurai, Yasushi},
title = {D-Tracker: Modeling Interest Diffusion in Social Activity Tensor Data Streams},
year = {2025},
booktitle = {SIGKDD},
pages = {460–471}
}

@inproceedings{10.1145/3580305.3599320,
author = {Xv, Guipeng and Lin, Chen and Guan, Wanxian and Gou, Jinping and Li, Xubin and Deng, Hongbo and Xu, Jian and Zheng, Bo},
title = {E-commerce Search via Content Collaborative Graph Neural Network},
year = {2023},
booktitle = {SIGKDD},
pages = {2885–2897}
}

@inproceedings{10.1145/3534678.3539213,
author = {Wen, Hongzhi and Ding, Jiayuan and Jin, Wei and Wang, Yiqi and Xie, Yuying and Tang, Jiliang},
title = {Graph Neural Networks for Multimodal Single-Cell Data Integration},
year = {2022},
booktitle = {SIGKDD},
pages = {4153–4163}
}

@inproceedings{zhu2025graphclip,
  title={Graphclip: Enhancing transferability in graph foundation models for text-attributed graphs},
  author={Zhu, Yun and Shi, Haizhou and Wang, Xiaotang and Liu, Yongchao and Wang, Yaoke and Peng, Boci and Hong, Chuntao and Tang, Siliang},
  booktitle={WWW},
  pages={2183--2197},
  year={2025}
}

@article{ha2016hypernetworks,
  title={Hypernetworks},
  author={Ha, David and Dai, Andrew and Le, Quoc V},
  journal={arXiv preprint arXiv:1609.09106},
  year={2016}
}

@incollection{taud2017multilayer,
  title={Multilayer perceptron (MLP)},
  author={Taud, Hind and Mas, Jean-Franccois},
  booktitle={Geomatic approaches for modeling land change scenarios},
  pages={451--455},
  year={2017}
  }

\newpage
\appendix
\raggedbottom
\section{Limitations}
Although \model\ improves transfer performance, it introduces additional inference overhead because graph evidence is updated across multiple reasoning steps. The current study also assumes that informative textual node attributes are available and evaluates a single graph backbone and a single foundation LLM, so the observed gains may not transfer uniformly to graphs with weaker text signals or to substantially different model families. Finally, while we evaluate eight public datasets across two domains and two task settings, broader validation on more graph types and larger-scale deployment conditions remains future work.

\section{Broader Impact}
This work may have positive impact by improving label-efficient reasoning on graph-structured data, which can support scientific discovery, recommendation, and knowledge-intensive decision support in low-resource settings. At the same time, stronger graph reasoning systems could be misused in high-stakes scenarios such as profiling, surveillance, or automated decisions over social or behavioral networks; accordingly, deployment should account for dataset bias, uncertainty, privacy, and application-specific oversight.

\section{Details of Contrastive Pre-training}
\label{Details of Contrastive Pre-training}

This appendix provides a detailed description of the contrastive pre-training objective used to learn transferable graph representations without relying on labeled data. The purpose of this stage is twofold: the graph encoder should produce stable node representations under stochastic graph perturbations, and the learned representation space should be compatible with the semantic geometry of LLM token embeddings.

Given an input graph $G=(V,E,X)$ with node features $X\in\mathbb{R}^{N\times d}$, we construct two stochastic views by perturbing both graph structure and node attributes. For structure augmentation, we randomly sample a binary masking matrix $\tilde{R}\in\{0,1\}^{N\times N}$ and apply it to the adjacency matrix via element-wise multiplication:
\begin{equation}
\tilde{A} = A \odot \tilde{R}.
\end{equation}
For attribute augmentation, a binary feature mask $\tilde{m}\in\{0,1\}^{d}$ is applied to each node feature:
\begin{equation}
\tilde{X} = [x_1\odot \tilde{m};\; x_2\odot \tilde{m};\; \dots;\; x_N\odot \tilde{m} ] .
\end{equation}
By applying these perturbations independently, we obtain two augmented views $(\tilde{X}_1,\tilde{A}_1)$ and $(\tilde{X}_2,\tilde{A}_2)$ of the same graph. These two views preserve the same underlying node identities and graph semantics, but expose the encoder to different local structures and feature observations.

The two augmented views are encoded by a shared graph encoder $f_{\text{GNN}}(\cdot)$:
\begin{equation}
U_* = f_{\text{GNN}}(\tilde{X}_*, \tilde{A}_*) \in \mathbb{R}^{N\times m}, 
\quad *\in\{1,2\},
\end{equation}
where $U_1$ and $U_2$ denote the node representation matrices of the two views, and $m$ is the output dimension of the graph encoder. For node $v_i$, we denote its representations in the two views as $u_i\in U_1$ and $u_i'\in U_2$.

We first encourage the representation of the same node to remain consistent across the two augmented views. Specifically, $u_i$ and $u_i'$ are treated as a positive pair, while the representations of other nodes from both views are used as negative samples. The contrastive objective for node $v_i$ is defined as:
\begin{equation}
\ell_{\text{ins}}(u_i,u_i') =
-\log
\frac{\exp(\theta(u_i,u_i')/\tau)}
{\sum\limits_{v \in \mathcal{N}(i)}
\exp(\theta(u_i,v)/\tau)} ,
\end{equation}
where $\mathcal{N}(i)=\{u_i'\}\cup\{u_j\}_{j\neq i}\cup\{u_j'\}_{j\neq i}$,
$\theta(\cdot,\cdot)$ denotes cosine similarity, and $\tau$ is the temperature parameter. The negative sign is used because the objective is minimized during training; therefore, minimizing this objective increases the similarity between the two representations of the same node and decreases the similarity to other nodes.

The node-level contrastive objective is computed symmetrically from both views and averaged over all nodes:
\begin{equation}
\mathcal{L}_{\text{ins}} =
\frac{1}{2N}\sum_{i=1}^{N}\bigl[
\ell_{\text{ins}}(u_i,u_i')+\ell_{\text{ins}}(u_i',u_i)
\bigr].
\end{equation}
This loss mainly preserves structural and semantic consistency at the node-instance level. However, it does not explicitly specify how the learned graph representation space should be organized with respect to the LLM embedding space.

To introduce such semantic guidance, we use the token embedding matrix of the LLM to construct a reference coordinate system. Concretely, we perform principal component analysis on the LLM token embeddings and retain the top-$P$ principal components:
\begin{equation}
C\in\mathbb{R}^{P\times m},
\end{equation}
where each row of $C$ corresponds to one principal direction of the LLM token embedding space. The output dimension of the graph encoder is set to $m$, matching the LLM token embedding dimension, so that graph representations can be projected onto these semantic directions. We then map the node representation matrices from the two augmented views into this LLM-induced coordinate system:
\begin{equation}
\tilde{U}_1 = U_1 C^{\top}, \quad
\tilde{U}_2 = U_2 C^{\top},
\end{equation}
where $\tilde{U}_1,\tilde{U}_2\in\mathbb{R}^{N\times P}$. In this formulation, the PCA projection is not an isolated post-processing step. Instead, it determines the space in which the following feature-level consistency constraint is applied.

After the projection, each column of $\tilde{U}_1$ or $\tilde{U}_2$ represents the responses of all nodes along one LLM-derived semantic direction. Let $\tilde{m}_i\in \tilde{U}_1^{\top}$ and $\tilde{n}_i\in \tilde{U}_2^{\top}$ denote the $i$-th projected feature vectors from the two views. We require the same semantic direction to be consistent across the two augmented views, while distinguishing it from other projected directions. The corresponding loss is defined as:
\begin{equation}
\mathcal{L}_{\text{fea}} =
-\frac{1}{P}\sum_{i=1}^{P}
\log
\frac{\exp(\theta(\tilde{m}_i,\tilde{n}_i)/\tau)}
{\sum_{j=1}^{P}\!\left[
\exp(\theta(\tilde{m}_i,\tilde{m}_j)/\tau)
+
\exp(\theta(\tilde{m}_i,\tilde{n}_j)/\tau)
\right]} .
\end{equation}
This loss is computed on the projected matrices $\tilde{U}_1$ and $\tilde{U}_2$, rather than directly on the original graph representation matrices $U_1$ and $U_2$. Therefore, the LLM token embedding space affects optimization through the PCA-induced projection, and the projected representations are explicitly involved in the feature-level contrastive objective.

The final pre-training objective combines the node-level consistency objective and the semantic-coordinate consistency objective:
\begin{equation}
\mathcal{L}_{\text{pre}} =
\frac{1}{2}\left(
\mathcal{L}_{\text{ins}} + \mathcal{L}_{\text{fea}}
\right).
\end{equation}
Through this objective, the graph encoder learns node representations that are stable under graph augmentations and are organized according to semantic directions derived from the LLM token embedding space. These pre-trained representations provide the initial graph evidence for subsequent graph-token construction and co-evolving reasoning.

\section{Full Prompt Templates}
\label{Full Prompt Templates}
In this section, we provide the full prompt templates used by our model during
multi-step reasoning and final prediction. For clarity, we use the Arxiv citation
graph as the representative example. In all templates, \texttt{<Node Token>}
denotes the graph-aware token representation injected into the language model.
The variable \texttt{\{title\}} denotes the textual title of the target paper,
and \texttt{\{categories\}} denotes the candidate label set for node
classification.

\subsection{Prompt Templates for Node Classification on Arxiv}
\label{app:prompt_nc_arxiv}

For node classification, the model is asked to classify a target paper into one
of the predefined Arxiv categories. During co-evolving chain-of-thought
reasoning, the prompt is dynamically constructed according to whether previous
reasoning states are available and whether the current step is the final
prediction step.

\paragraph{CoT first step.}
When no previous reasoning state is available and the current step is not the
final prediction step, we use the following prompt.

\begin{promptbox}{Node Classification Prompt: CoT First Step}
\small
Given a node \texttt{<Node Token>} in a paper citation graph.

Title: \texttt{\{title\}}

Please classify this paper into one of the following categories:

\texttt{\{categories\}}

Provide your reasoning step by step. Give your next step of reasoning only, do NOT give the final answer yet.

Your reasoning:
\end{promptbox}

\paragraph{CoT intermediate step.}
When previous reasoning states have been accumulated but the current step is
still not the final prediction step, the model is prompted to continue the
reasoning process based on the previous thoughts.

\begin{promptbox}{Node Classification Prompt: CoT Intermediate Step}
\small
Given a node \texttt{<Node Token>} in a paper citation graph.

Title: \texttt{\{title\}}

Please classify this paper into one of the following categories:

\texttt{\{categories\}}

Your previous reasoning:

\texttt{\{previous\_reasoning\}}

Continue your analysis. Provide your next step of reasoning only, do NOT give the final answer yet.

Your next reasoning step:
\end{promptbox}

\paragraph{Final prediction with reasoning states.}
When previous reasoning states are available and the model reaches the final
prediction step, the prompt asks the model to produce the final answer based on
all accumulated reasoning states.

\begin{promptbox}{Node Classification Prompt: Final Prediction with Reasoning}
\small
Given a node \texttt{<Node Token>} in a paper citation graph.

Title: \texttt{\{title\}}

Please classify this paper into one of the following categories:

\texttt{\{categories\}}

Your previous reasoning:

\texttt{\{previous\_reasoning\}}

Based on all the reasoning above, provide your final answer.

Answer:
\end{promptbox}

\paragraph{Final prediction without reasoning states.}
In the degenerated case where no intermediate reasoning state is available but
the model is required to directly produce the final answer, we use the following
prompt.

\begin{promptbox}{Node Classification Prompt: Direct Final Prediction}
\small
Given a node \texttt{<Node Token>} in a paper citation graph.

Title: \texttt{\{title\}}

Please classify this paper into one of the following categories:

\texttt{\{categories\}}

Provide your final answer.
\end{promptbox}

\subsection{Prompt Templates for Link Prediction on Arxiv}
\label{app:prompt_lp_arxiv}

For link prediction, the model is given two papers from the Arxiv citation graph
and is asked to determine whether there exists a citation link between them. To
ensure a clean binary prediction, we append an explicit instruction requiring the
model to answer with only \texttt{yes} or \texttt{no} at the final prediction
stage.

\paragraph{CoT first step.}
When no previous reasoning state is available and the current step is not the
final prediction step, the model is prompted to generate only the next reasoning
step.

\begin{promptbox}{Link Prediction Prompt: CoT First Step}
\small
Given two nodes \texttt{<Node Token$^{A}$>} and \texttt{<Node Token$^{B}$>} in a paper citation graph.

Title of node A: \texttt{\{title\_a\}}

Title of node B: \texttt{\{title\_b\}}

Question: Is there a citation link between these two papers?

Please answer with only `yes' or `no'. Do not answer your thought.

Provide your reasoning step by step. Give your next step of reasoning only, do NOT give the final answer yet.

Your reasoning:
\end{promptbox}

\paragraph{CoT intermediate step.}
When previous reasoning states are available but the current step is not the
final prediction step, the model continues its analysis conditioned on the
accumulated reasoning history.

\begin{promptbox}{Link Prediction Prompt: CoT Intermediate Step}
\small
Given two nodes \texttt{<Node Token$^{A}$>} and \texttt{<Node Token$^{B}$>} in a paper citation graph.

Title of node A: \texttt{\{title\_a\}}

Title of node B: \texttt{\{title\_b\}}

Question: Is there a citation link between these two papers?

Please answer with only `yes' or `no'. Do not answer your thought.

Your previous reasoning:

\texttt{\{previous\_reasoning\}}

Continue your analysis. Provide your next step of reasoning only, do NOT give the final answer yet.

Your next reasoning step:
\end{promptbox}

\paragraph{Final prediction with reasoning states.}
When the model reaches the final prediction step, it is required to output a
binary answer based on all previous reasoning states.

\begin{promptbox}{Link Prediction Prompt: Final Prediction with Reasoning}
\small
Given two nodes \texttt{<Node Token$^{A}$>} and \texttt{<Node Token$^{B}$>} in a paper citation graph.

Title of node A: \texttt{\{title\_a\}}

Title of node B: \texttt{\{title\_b\}}

Question: Is there a citation link between these two papers?

Please answer with only `yes' or `no'. Do not answer your thought.

Your previous reasoning:

\texttt{\{previous\_reasoning\}}

Based on all the reasoning above, provide your final answer.

Answer:
\end{promptbox}

\paragraph{Final prediction without reasoning states.}
When no intermediate reasoning state is available, the model directly predicts
whether a citation link exists between the two papers.

\begin{promptbox}{Link Prediction Prompt: Direct Final Prediction}
\small
Given two nodes \texttt{<Node Token$^{A}$>} and \texttt{<Node Token$^{B}$>} in a paper citation graph.

Title of node A: \texttt{\{title\_a\}}

Title of node B: \texttt{\{title\_b\}}

Question: Is there a citation link between these two papers?

Please answer with only `yes' or `no'. Do not answer your thought.

Provide your final answer.
\end{promptbox}

\section{Algorithm and Complexity Analysis}
\label{Algorithm and Complexity Analysis}

\begin{algorithm}[h]
\caption{\textsc{Co-Evolving Chain-of-Thought Prompting}}
\label{alg:coevot}
\begin{algorithmic}[1]
\State \textbf{Input:} Graph $G=(V,E,X)$, node texts $\mathcal{S}$, task question $\mathcal{Q}$, labeled source data $\mathcal{D}_s$, frozen graph encoder $\mathrm{GE}(\cdot;\Theta_0)$, frozen LLM $\mathcal{M}$, reasoning steps $K$, update weight $\alpha$
\State \textbf{Output:} Optimized trainable parameter set $\Phi=\{\phi,\psi\}$
\State Initialize the linear projector parameter $\phi$
\State Initialize the condition network parameter $\psi$
\While{not converged}
    \State Sample a mini-batch $\mathcal{B}\subset\mathcal{D}_s$
    \State $H^0 \leftarrow \mathrm{GE}(X,G;\Theta_0)$
    \State $H_{\mathrm{token}}^0 \leftarrow \mathrm{Linear}(H^0;\phi)$
    \For{each node $v\in\mathcal{B}$}
        \State $\vec{h}_v^0 \leftarrow H_v^0$
        \State $\vec{h}_{\mathrm{token},v}^0 \leftarrow H_{\mathrm{token},v}^0$
        \State $\mathcal{R}_v^0 \leftarrow \emptyset$
    \EndFor
    \For{$k=1$ to $K$}
        \For{each node $v\in\mathcal{B}$}
            \State $\langle\mathrm{Instruction}\rangle_v^k \leftarrow \mathrm{BuildInstruction}(\vec{h}_{\mathrm{token},v}^{k-1},\mathcal{S}_v)$
            \State $q_v^k \leftarrow \mathrm{BuildPrompt}(\langle\mathrm{Instruction}\rangle_v^k,\mathcal{R}_v^{k-1},\mathcal{Q})$
            \State $\mathrm{Thought}_v^k \leftarrow \mathcal{M}(q_v^k)$
            \State $\vec{t}_v^k \leftarrow \mathrm{Rep}_{\mathcal{M}}(\mathrm{Thought}_v^k)$
            \State $\mathcal{R}_v^k \leftarrow \mathcal{R}_v^{k-1}\cup\{\mathrm{Thought}_v^k\}$
            \State $\vec{p}_v^k \leftarrow \mathrm{CondNet}(\vec{t}_v^1\oplus\cdots\oplus\vec{t}_v^k;\psi)$
            \State $\vec{h}_v^{k+1} \leftarrow \vec{h}_v^{k}+\alpha\cdot\vec{p}_v^k$
            \State $\vec{h}_{\mathrm{token},v}^k \leftarrow \mathrm{Linear}(\vec{h}_v^k;\phi)$
        \EndFor
    \EndFor
    \For{each $(v,y_v)\in\mathcal{B}$}
        \State $\langle\mathrm{Instruction}\rangle_v^K \leftarrow \mathrm{BuildInstruction}(\vec{h}_{\mathrm{token},v}^{K},\mathcal{S}_v)$
        \State $q_v^{\mathrm{ans}} \leftarrow \mathrm{BuildAnswerPrompt}(\langle\mathrm{Instruction}\rangle_v^K,\mathcal{R}_v^K,\mathcal{Q})$
    \EndFor
    \State $\mathcal{L}_{\mathrm{down}}(\phi,\psi) \leftarrow -\sum_{(v,y_v)\in\mathcal{B}}\log p_{\mathcal{M}}(y_v\mid q_v^{\mathrm{ans}};\phi,\psi)$
    \State Update $\phi$ and $\psi$ by backpropagating $\mathcal{L}_{\mathrm{down}}(\phi,\psi)$
\EndWhile
\State $\Phi \leftarrow \{\phi,\psi\}$
\State \Return $\Phi$
\end{algorithmic}
\end{algorithm}

\noindent \textbf{Algorithm.}
We outline the overall workflow of \method{COEVOT} in Algorithm~\ref{alg:coevot}. 
We first initialize the trainable linear projector parameter $\phi$ and the condition network parameter $\psi$, and denote the trainable parameter set as $\Phi=\{\phi,\psi\}$. 
At each optimization iteration, we sample a mini-batch $\mathcal{B}\subset\mathcal{D}_s$ from the labeled source data. 
The frozen pre-trained graph encoder is applied to the input graph to obtain the initial graph embeddings, which are then mapped into the LLM token space by the linear projector. 
For each node in the mini-batch, we initialize its node-level graph embedding, graph-token representation, and an empty thought buffer for storing the intermediate natural-language thoughts generated during reasoning.

The model then performs $K$ rounds of co-evolving reasoning. 
At each step, the current graph token and node text are first used to construct a node-specific instruction, which is further combined with the task question and the accumulated thought buffer to form the full prompt. 
The frozen LLM generates an intermediate thought conditioned on this prompt. 
We then obtain the representation of this thought in the LLM embedding space and append the generated thought to the buffer. 
The condition network takes the accumulated thought representations as conditioning signals and produces a node-specific conditional prompt, which is used to rewrite the current node embedding through the residual update defined in Sect.~\ref{Text-to-Graph Token Rewriting}. 
The updated node embedding is projected again into the LLM token space and serves as the graph-token evidence for the next reasoning step. 
Through this alternating process, graph-to-text reasoning and text-to-graph rewriting form a closed feedback loop, enabling the graph evidence and the reasoning trajectory to evolve jointly.

After completing $K$ reasoning steps, we construct the final answer prompt using the final graph token, node text, task question, and the complete thought buffer. 
The frozen LLM then produces the predictive distribution for the downstream task. 
The supervised objective is computed over the mini-batch, and only $\phi$ and $\psi$ are updated by backpropagation, while both the graph encoder and the LLM remain frozen throughout adaptation. 
The procedure terminates when training converges and returns the optimized trainable parameter set $\Phi=\{\phi,\psi\}$.

\noindent \textbf{Complexity analysis.}
Given a downstream graph $G=(\mathcal{V},\mathcal{E})$, each optimization iteration begins with one forward pass of the frozen graph encoder. 
For an $L_g$-layer message-passing graph encoder, the cost of computing node representations is $O(L_g|\mathcal{E}|)$, which can be approximated as $O(L_g\bar{d}|\mathcal{V}|)$ when each node aggregates at most $\bar{d}$ neighbors on average. 
Here, $|\mathcal{V}|$ and $|\mathcal{E}|$ denote the numbers of nodes and edges, respectively.

After graph encoding, the model performs $K$ rounds of LLM-based reasoning for a mini-batch of size $B$. 
Let $T$ denote the prompt sequence length and $d_{\mathcal{M}}$ denote the hidden dimension of the LLM. 
The dominant cost of multi-step reasoning is the LLM inference cost, which is $O(BKT^2d_{\mathcal{M}})$, where the final answer generation can be absorbed into the same order term. 
At each reasoning step, the condition network maps the accumulated thought representations into node-specific conditional prompts, followed by residual graph-embedding rewriting and linear graph-token projection. 
These operations are lightweight compared with LLM inference and are treated as lower-order overhead. Therefore, the overall cost of one optimization iteration is $O\bigl(L_g|\mathcal{E}| + BKT^2d_{\mathcal{M}}\bigr).$

In practice, the LLM inference term usually dominates the total runtime, and the computational cost is mainly governed by the mini-batch size $B$, the number of reasoning steps $K$, and the prompt sequence length $T$.

\section{Implementation Details}
\label{Implementation Details}
We directly adopt the datasets and data splits provided by \method{TEA-GLM}~\citep{wang2024llms} for all experiments without any modification. This ensures that all methods are evaluated on exactly the same data and split configurations, enabling strictly controlled and fair comparisons. We report Accuracy and Macro-F1 for node classification, and AUC for link prediction.

\stitle{\model\ setting.} We adopt a 2-layer GraphSAGE as the graph encoder and initialize it with the pre-trained checkpoint released by \method{TEA-GLM}~\citep{wang2024llms}, which is trained under an identical pre-training protocol. This setup ensures that performance differences arise from the proposed method rather than discrepancies in graph representation learning.
The model is trained for 60 epochs with a batch size of 512 using the Adam optimizer and a learning rate of $2 \times 10^{-2}$. The linear projector is trained separately for 2 epochs with a batch size of 2 and a learning rate of $1 \times 10^{-3}$. We set the number of reasoning steps to $K=2$ and the condition weight to $\alpha=0.4$. This configuration provides a favorable trade-off between performance and computational cost, and is consistently used across all experiments.

\stitle{\method{GCoT} setting.} 
For \method{GCoT}, we use a 3-layer GCN as the backbone, with the hidden dimension set to $256$. 
The number of inference steps is set to $2$. The condition-net is implemented as a two-layer MLP with a bottleneck architecture, where the input dimension is $256$ and the hidden dimension is set to $32$.

\stitle{\method{GOFA} setting.}
For \method{GOFA}, we set the optimizer to AdamW, with a learning rate of $1 \times 10^{-4}$, weight decay of $0.1$, batch size of $8$, dropout of $0.0$, gradient clipping of $0.5$, gradient accumulation of $8$, and a maximum LLM sequence length of $128$.

\stitle{Environment.} All experiments were conducted on Ubuntu 22.04.2, using a machine equipped with a single NVIDIA vGPU with 48\,GB memory.

\section{Further Descriptions of Datasets}
\label{Further Descriptions of Datasets}
We provide more comprehensive descriptions of the datasets\footnote{\url{https://ogb.stanford.edu/}}
\footnote{\url{https://github.com/XiaoxinHe/TAPE}}
\footnote{\url{https://github.com/WenZhihao666/G2P2}}
\footnote{\url{https://github.com/sktsherlock/TAG-Benchmark}} used in our experiments. These datasets are summarized in Table ~\ref{tab:dataset_stats_domain}.

\begin{table}[h]
  \centering
  \caption{ Summary of datasets.}
  \label{tab:dataset_stats_domain}
  \setlength{\tabcolsep}{4pt}
  \renewcommand{\arraystretch}{1.0}

  \begin{tabular}{llrrr}
    \toprule
    Domain & Dataset & Nodes & Edges & Classes \\
    \midrule
    \multirow{3}{*}{Citation} 
      & Arxiv   & 169,343 & 1,166,243 & 40 \\
      & Pubmed  & 19,717  & 44,338    & 3  \\
      & Cora    & 25,120  & 91,140    & 70 \\
    \midrule
    \multirow{5}{*}{E-commerce} 
      & Computer   & 87,229  & 721,081   & 10 \\
      & Photo      & 48,362  & 500,928   & 12 \\
      & Children  & 76,875  & 1,554,578 & 24 \\
      & History   & 41,551  & 358,574   & 12 \\
      & Sports & 173,055 & 1,773,500 & 13 \\
    \bottomrule
  \end{tabular}
\end{table}

\noindent (1) \textbf{Citation Datasets.}

\textbf{Arxiv}~\citep{hu2020open} is a directed citation network constructed from computer science papers collected from the arXiv preprint repository. Nodes represent individual papers, and directed edges indicate citation relationships.

\textbf{PubMed}~\citep{he2023harnessing} contains 19,717 biomedical publications related to diabetes. Each paper is categorized into one of three classes: experimentally induced diabetes, type~1 diabetes, or type~2 diabetes, according to its primary research focus.

\textbf{Cora}~\citep{wen2023augmenting} provides a citation network for analyzing machine learning research papers. We use an extended version that offers more fine-grained category annotations.

\noindent (2) \textbf{E-commerce Datasets}.

\noindent All e-commerce datasets are collected from a unified benchmark~\citep{yan2023comprehensive}.

\textbf{Books-Children} and \textbf{Books-History} are derived from the Amazon Books domain, containing items labeled under the second-level categories ``Children'' and ``History'', respectively. The labels correspond to third-level book categories.

\textbf{Ele-Computers} and \textbf{Ele-Photo} are extracted from the electronics domain, where items are associated with the second-level categories ``Computers'' and ``Photo'' and classified at a finer granularity.

\textbf{Sports-Fitness} contains fitness-related products. In this dataset, nodes denote items, and edges indicate frequent co-purchase or co-view relationships between items.

\section{Further Descriptions of Baselines}
\label{Further Descriptions of Baselines}
We provide additional details about the baseline methods used in our experiments.

\noindent (1) GNN as predictor.

\method{\textbf{GCN}}~\citep{kipf2016semi}: A graph neural network that propagates and aggregates node features via normalized neighborhood averaging, enabling nodes to capture local structural and attribute information.

\method{\textbf{GraphSAGE}}~\citep{hamilton2017inductive}: An inductive graph representation learning framework that samples and aggregates features from a node's local neighborhood, allowing generalization to unseen nodes and graphs.

\method{\textbf{GAT}}~\citep{velivckovic2017graph}: A graph neural network that employs self-attention mechanisms to assign adaptive importance weights to neighboring nodes, thereby refining neighborhood aggregation.

\method{\textbf{DGI}}~\citep{velivckovic2018deep}: A self-supervised graph pre-training method that maximizes mutual information between local node representations and a global graph summary to enhance structural awareness.

\method{\textbf{GKD}}~\citep{yang2022geometric}: A knowledge distillation framework for graphs that transfers structural and semantic information from a teacher model to a lightweight student GNN to improve performance efficiency.

\method{\textbf{GLNN}}~\citep{eliasof2024global}: A graph learning framework that integrates global structural information into node-level representations, aiming to alleviate the locality bias of standard message-passing GNNs.

\method{\textbf{NodeFormer}}~\citep{wu2022nodeformer}: A graph Transformer model that replaces message passing with kernelized global attention, enabling scalable modeling of long-range node dependencies.

\method{\textbf{DIFFormer}}~\citep{wu2023difformer}: A diffusion-based graph Transformer that incorporates graph diffusion processes into attention computation to capture both local and global structural patterns.

\method{\textbf{OFA}}~\citep{liu2023one}: A unified graph foundation model that reformulates diverse graph tasks into a single optimization framework, enabling task-agnostic and transfer-friendly graph representation learning.

\method{\textbf{GCoT}}~\citep{yu2025gcot}: A graph reasoning framework that introduces chain-of-thought style intermediate representations to iteratively refine node embeddings and improve zero-shot generalization.

\noindent (2) LLM as predictor.

\method{\textbf{Vicuna}}~\citep{team2023vicuna}: A chat-oriented large language model fine-tuned from LLaMA using high-quality conversational data to enhance instruction-following and dialogue capabilities.

\method{\textbf{GraphGPT}}~\citep{tang2024graphgpt}: A graph-aware large language model that encodes graph structures into textual representations, enabling large language models to perform graph understanding and reasoning.

\method{\textbf{LLaGA}}~\citep{chen2024llaga}: A graph-language alignment framework that bridges graph representations and large language models through instruction-guided adaptation for graph learning tasks.

\method{\textbf{TEA-GLM}}~\citep{wang2024llms}: A task-enhanced graph-language model that aligns graph embeddings with language model representations to support zero-shot and transfer learning on graph tasks.

\method{\textbf{GOFA}}~\citep{kong2024gofa}: A graph-oriented foundation agent that leverages large language models 

\section{Macro-F1 on node classification task}
\label{F1 score on node classification task}

\begin{table}[h]
  \caption{Macrof-F1 results on node classification.}
  \label{tab:f1}
  \centering
  \small
  \setlength{\tabcolsep}{2.5pt}
  \renewcommand{\arraystretch}{1.2}
  \resizebox{\columnwidth}{!}{
  \begin{tabular}{llcccccc}
    \toprule
    \multirow{2}{*}{Model type} & \multirow{2}{*}{Model} &
    \multicolumn{2}{c}{Citation} & \multicolumn{4}{c}{E-commerce} \\
    \cmidrule(lr){3-4}\cmidrule(lr){5-8}
    & & Pubmed & Cora & Children & History & Photo & Sports \\
    \midrule

    & \method{MLP} & 0.246$\pm$0.042 & 0.009$\pm$0.004 & 0.007$\pm$0.007 & 0.023$\pm$0.008 & 0.041$\pm$0.023 & 0.019$\pm$0.005 \\
    \midrule

    \multirow{10}{*}{GNN as predictor}
    & \method{GCN}        & 0.187$\pm$0.021 & 0.007$\pm$0.001 & 0.006$\pm$0.004 & 0.024$\pm$0.013 & 0.034$\pm$0.007 & 0.017$\pm$0.009 \\
    & \method{GraphSAGE}  & 0.257$\pm$0.084 & 0.007$\pm$0.003 & 0.005$\pm$0.003 & 0.029$\pm$0.024 & 0.020$\pm$0.011 & 0.021$\pm$0.004 \\
    & \method{GAT}        & 0.259$\pm$0.065 & 0.006$\pm$0.001 & 0.063$\pm$0.067 & 0.159$\pm$0.117 & 0.036$\pm$0.035 & 0.091$\pm$0.090 \\
    & \method{DGI}        & 0.213$\pm$0.127 & 0.004$\pm$0.002 & 0.012$\pm$0.004 & 0.038$\pm$0.015 & 0.045$\pm$0.015 & 0.018$\pm$0.005 \\
    & \method{GKD}        & 0.247$\pm$0.039 & 0.004$\pm$0.001 & 0.028$\pm$0.003 & 0.060$\pm$0.008 & 0.049$\pm$0.015 & 0.050$\pm$0.008 \\
    & \method{GLNN}       & 0.221$\pm$0.033 & 0.006$\pm$0.001 & 0.021$\pm$0.003 & 0.064$\pm$0.007 & 0.057$\pm$0.002 & 0.052$\pm$0.003 \\
    & \method{NodeFormer} & 0.232$\pm$0.089 & 0.008$\pm$0.003 & 0.019$\pm$0.008 & 0.046$\pm$0.031 & 0.055$\pm$0.006 & 0.049$\pm$0.009 \\
    & \method{DIFFormer}  & 0.187$\pm$0.007 & 0.007$\pm$0.002 & 0.002$\pm$0.002 & 0.050$\pm$0.019 & 0.069$\pm$0.010 & 0.045$\pm$0.007 \\
    & \method{OFA}        & 0.287$\pm$0.059 & 0.091$\pm$0.013 & 0.017$\pm$0.010 & 0.026$\pm$0.007 & 0.103$\pm$0.007 & 0.043$\pm$0.021 \\
    & \method{GCoT}       & 0.257$\pm$0.015 & 0.025$\pm$0.009 & 0.043$\pm$0.005 & 0.032$\pm$0.014 & 0.098$\pm$0.014 & 0.074$\pm$0.018 \\
    \midrule

    \multirow{7}{*}{LLM as predictor}
    & \method{Vicuna-7B-v1.5} & 0.629$\pm$0.024 & 0.109$\pm$0.002 & 0.179$\pm$0.002 & 0.349$\pm$0.003 & 0.383$\pm$0.001 & 0.410$\pm$0.002 \\
    & \method{GraphGPT-std}   & 0.649 & 0.082 & -- & -- & -- & -- \\
    & \method{GraphGPT-cot}   & 0.482 & 0.127 & -- & -- & -- & -- \\
    & \method{LLaGA}          & 0.778$\pm$0.056 & 0.108$\pm$0.014 & 0.163$\pm$0.029 & 0.144$\pm$0.025 & 0.362$\pm$0.039 & \underline{0.446$\pm$0.035} \\
    & \method{TEA-GLM}        & \underline{0.839$\pm$0.012} & \underline{0.148$\pm$0.015} & \underline{0.252$\pm$0.005} & 0.365$\pm$0.011 & 0.421$\pm$0.032 & 0.430$\pm$0.009 \\
    & \method{GOFA}           & 0.826$\pm$0.032 & 0.146$\pm$0.018 & 0.185$\pm$0.012 & \underline{0.379$\pm$0.017} & \underline{0.442$\pm$0.027} & 0.415$\pm$0.006 \\
    & \model                  & \textbf{0.879$\pm$0.014} & \textbf{0.162$\pm$0.016} & \textbf{0.255$\pm$0.008} & \textbf{0.395$\pm$0.028} & \textbf{0.458$\pm$0.018} & \textbf{0.462$\pm$0.013} \\
    \bottomrule
  \end{tabular}}
\end{table}

Tables~\ref{tab:f1} further reports the cross-dataset zero-shot node classification results in terms of Macro-F1. We observe that \model\ consistently achieves the best performance on all six target datasets, with particularly clear improvements over strong graph--LLM baselines such as TEA-GLM and GOFA. Since Macro-F1 is more sensitive to class imbalance and fine-grained discrimination, these results provide stronger evidence that \model\ learns more transferable and discriminative representations under distribution shift. In contrast, traditional GNN-based methods perform poorly in this setting, reflecting their limited ability to generalize across graphs with shifted structures and semantics. Although pure LLM-based methods exhibit relatively stronger robustness due to their semantic prior knowledge, they remain inferior to \model\ because they do not explicitly refine structural evidence during reasoning. By dynamically coupling graph-to-text guided reasoning with text-to-graph token rewriting, \model\ enables intermediate reasoning signals to iteratively reshape graph representations, which leads to more balanced and reliable classification across categories.

\section{Visualization}
\label{Visualization}

\begin{figure}[h]
    \centering
    \includegraphics[width=0.7\textwidth]{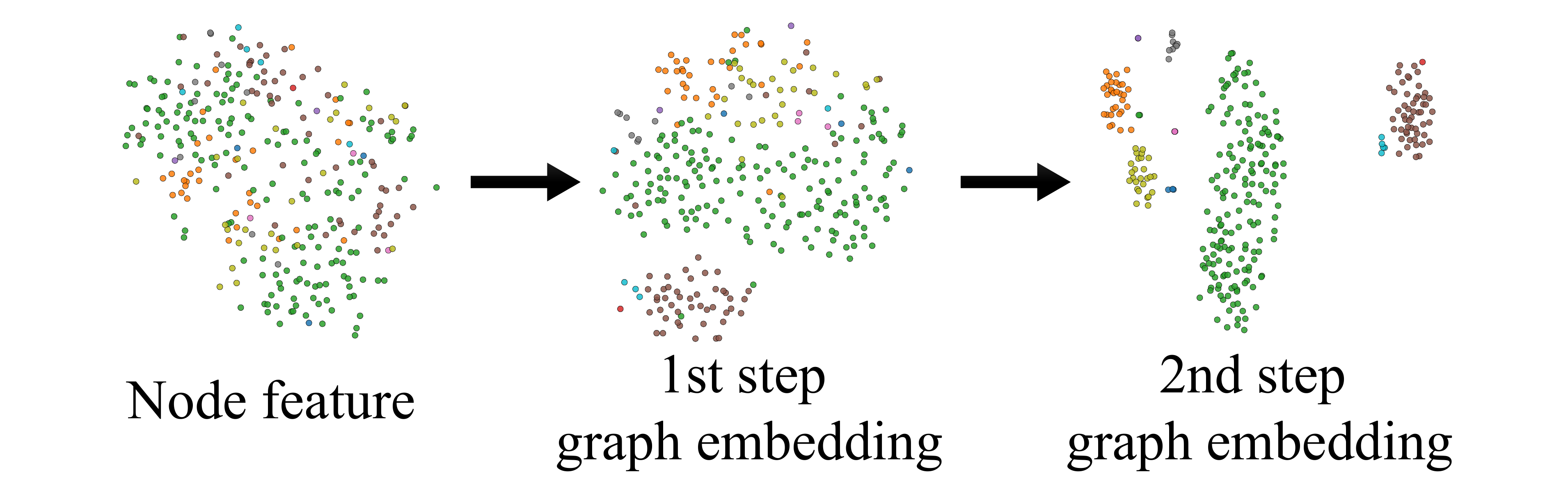}
    \caption{Visualization of step-wise node representation evolution on History.}
    \label{fig:cot_vis_b}
\end{figure}

The results on History show a similar trend to ArXiv. As shown in Fig.~\ref{fig:cot_vis_b}, the initial node features are highly entangled across classes, suggesting limited class separability under distribution shift. After the first reasoning step, the embeddings begin to form coarse clusters, indicating that intermediate reasoning introduces task-relevant semantic and structural signals. After the second reasoning step, the clusters become more compact and clearly separated, showing that Text-to-Graph Token Rewriting progressively reshapes the embedding space and leads to more discriminative representations.

\section{Computational Efficiency Analysis}

We evaluate the computational efficiency of \model\ through zero-shot node classification, with inference time comparisons reported in Table~\ref{tab:efficiency}.
\model\ incurs higher inference latency compared to existing baselines such as \method{TEA-GLM} and \method{GOFA}. This overhead primarily stems from the multi-step reasoning process, where graph representations are iteratively updated through Text-to-Graph Token Rewriting at each step.
Notably, the inference cost increases approximately linearly with the number of reasoning steps, indicating a predictable and controllable computational overhead. In practice, we find that a small number of steps (e.g., $K=2$) is sufficient to achieve strong performance, which helps limit the overall cost.
Despite the additional computation, \model\ consistently delivers substantial performance gains (Tables~\ref{tab:llm_for_graph_predictor} and~\ref{tab:f1}), suggesting a favorable trade-off between accuracy and efficiency. These results indicate that the proposed reasoning–structure coupling mechanism improves generalization at the cost of moderate and controllable inference overhead.

\begin{table}[h]
\centering
\caption{Inference time (s) comparison under zero-shot node classification.}
\label{tab:efficiency}
\renewcommand{\arraystretch}{1}
\begin{tabular}{l|rrrr}
\hline
Methods & Pubmed & Cora & Children & History \\
\hline
\method{GOFA} & 946.1 & 1274.1 & 539.9 & 481.3 \\
\method{TEA-GLM}        & 1749.4 & 2535.8 & 1120.3 & 1092.6 \\
\model        & 6732.9 & 4820.6 & 2945.4 & 3957.2 \\
\hline
\end{tabular}
\end{table}

\newpage

\section*{NeurIPS Paper Checklist}

\begin{enumerate}

\item {\bf Claims}
    \item[] Question: Do the main claims made in the abstract and introduction accurately reflect the paper's contributions and scope?
    \item[] Answer: \answerYes{}
    \item[] Justification: The abstract and introduction summarize the proposed co-evolving CoT framework, its core components, and the empirical improvements demonstrated in the experiments; these claims are aligned with the method and results sections.

\item {\bf Limitations}
    \item[] Question: Does the paper discuss the limitations of the work performed by the authors?
    \item[] Answer: \answerYes{}
    \item[] Justification: The paper includes a short limitations discussion in the main text, covering computational overhead, dependence on textual node attributes, and the current evaluation scope.

\item {\bf Theory assumptions and proofs}
    \item[] Question: For each theoretical result, does the paper provide the full set of assumptions and a complete (and correct) proof?
    \item[] Answer: \answerNA{}
    \item[] Justification: The paper is primarily empirical and methodological; it does not present formal theorems or proofs.

    \item {\bf Experimental result reproducibility}
    \item[] Question: Does the paper fully disclose all the information needed to reproduce the main experimental results of the paper to the extent that it affects the main claims and/or conclusions of the paper (regardless of whether the code and data are provided or not)?
    \item[] Answer: \answerYes{}
    \item[] Justification: The paper specifies the backbone models, datasets, training/evaluation protocol, hyperparameter analyses, and an anonymized code repository for review.

\item {\bf Open access to data and code}
    \item[] Question: Does the paper provide open access to the data and code, with sufficient instructions to faithfully reproduce the main experimental results, as described in supplemental material?
    \item[] Answer: \answerYes{}
    \item[] Justification: The submission provides an anonymized repository URL in the paper for review. The datasets used are public benchmarks and are cited in the paper and appendix.

\item {\bf Experimental setting/details}
    \item[] Question: Does the paper specify all the training and test details (e.g., data splits, hyperparameters, how they were chosen, type of optimizer) necessary to understand the results?
    \item[] Answer: \answerYes{}
    \item[] Justification: The paper describes the training protocol, zero-shot evaluation settings, model backbones, task settings, and reports additional analyses of key hyperparameters and computational efficiency.

\item {\bf Experiment statistical significance}
    \item[] Question: Does the paper report error bars suitably and correctly defined or other appropriate information about the statistical significance of the experiments?
    \item[] Answer: \answerYes{}
    \item[] Justification: The main node classification results report mean and standard deviation when repeated runs are available. Link prediction and ablation studies report single-run or deterministic results under the same evaluation protocol, with consistent metrics across all methods and datasets.


\item {\bf Experiments compute resources}
    \item[] Question: For each experiment, does the paper provide sufficient information on the computer resources (type of compute workers, memory, time of execution) needed to reproduce the experiments?
    \item[] Answer: \answerNo{}
    \item[] Justification: The paper discusses computational efficiency and relative inference cost, but does not yet give a full hardware specification or end-to-end training-time breakdown for all experiments.
    
\item {\bf Code of ethics}
    \item[] Question: Does the research conducted in the paper conform, in every respect, with the NeurIPS Code of Ethics \url{https://neurips.cc/public/EthicsGuidelines}?
    \item[] Answer: \answerYes{}
    \item[] Justification: The work uses standard public benchmarks and anonymized review materials, and we are not aware of any aspect of the research that conflicts with the NeurIPS Code of Ethics.

\item {\bf Broader impacts}
    \item[] Question: Does the paper discuss both potential positive societal impacts and negative societal impacts of the work performed?
    \item[] Answer: \answerYes{}
    \item[] Justification: The paper includes a brief broader-impact discussion in the appendix, noting both the potential benefit of improved graph reasoning and possible misuse risks in high-stakes applications.
    
\item {\bf Safeguards}
    \item[] Question: Does the paper describe safeguards that have been put in place for responsible release of data or models that have a high risk for misuse (e.g., pre-trained language models, image generators, or scraped datasets)?
    \item[] Answer: \answerNA{}
    \item[] Justification: The paper does not release a new high-risk generative model or scraped dataset; it builds on existing public assets and benchmark datasets.

\item {\bf Licenses for existing assets}
    \item[] Question: Are the creators or original owners of assets (e.g., code, data, models), used in the paper, properly credited and are the license and terms of use explicitly mentioned and properly respected?
    \item[] Answer: \answerNo{}
    \item[] Justification: The paper cites the original datasets and baseline methods, but it does not yet systematically enumerate license names and terms of use for every external asset in the current draft.

\item {\bf New assets}
    \item[] Question: Are new assets introduced in the paper well documented and is the documentation provided alongside the assets?
    \item[] Answer: \answerYes{}
    \item[] Justification: The paper indicates that code is released through an anonymized repository for review; the methodological description and appendix provide the information needed to document the released implementation.

\item {\bf Crowdsourcing and research with human subjects}
    \item[] Question: For crowdsourcing experiments and research with human subjects, does the paper include the full text of instructions given to participants and screenshots, if applicable, as well as details about compensation (if any)? 
    \item[] Answer: \answerNA{}
    \item[] Justification: The paper does not involve crowdsourcing or research with human subjects.

\item {\bf Institutional review board (IRB) approvals or equivalent for research with human subjects}
    \item[] Question: Does the paper describe potential risks incurred by study participants, whether such risks were disclosed to the subjects, and whether Institutional Review Board (IRB) approvals (or an equivalent approval/review based on the requirements of your country or institution) were obtained?
    \item[] Answer: \answerNA{}
    \item[] Justification: The paper does not involve human-subjects research.

\item {\bf Declaration of LLM usage}
    \item[] Question: Does the paper describe the usage of LLMs if it is an important, original, or non-standard component of the core methods in this research? Note that if the LLM is used only for writing, editing, or formatting purposes and does \emph{not} impact the core methodology, scientific rigor, or originality of the research, declaration is not required.
    \item[] Answer: \answerYes{}
    \item[] Justification: LLM usage is central to the proposed method and is explicitly described throughout the method section, adaptation/evaluation setup, and experiments.

\end{enumerate}

\end{document}